\pdfoutput=1

\documentclass[11pt]{article}
\usepackage{acl}

\usepackage{times}
\usepackage{latexsym}
\usepackage[T1]{fontenc}
\usepackage[utf8]{inputenc}
\usepackage{microtype}
\usepackage{inconsolata}
\usepackage{graphicx}
\usepackage{amsmath}
\usepackage{booktabs}
\usepackage{lipsum}
\usepackage[caption=false]{subfig}
\usepackage{float}
\usepackage{multirow}
\usepackage{amsmath}
\usepackage{changepage}
\usepackage{mathtools}
\usepackage{breqn}
\usepackage{stfloats}

\DeclareMathOperator*{\argmax}{arg\,max}

\usepackage[disable]{todonotes}                                 
\makeatletter
\newcommand*\iftodonotes{\if@todonotes@disabled\expandafter\@secondoftwo\else\expandafter\@firstoftwo\fi}  
\makeatother

\newcommand{\tinytodo}[1]{\todo[size=\tiny]{#1}}
\newcommand{\smalltodo}[1]{\todo[size=\small]{#1}}

\newcommand{\pragoracle}{\texttt{Oracle}}
\newcommand{\pragrandom}{\texttt{Random}}
\newcommand{\pragans}{\texttt{Answer-Rec}}
\newcommand{\pragsource}{\texttt{Source-Rec}}
\newcommand{\pragas}{\texttt{Ans-Src-Rec}}

\newcommand{\literalsumm}{\texttt{Literal-Summ-Only}}
\newcommand{\pragansonly}{\texttt{Answer-Rec-Only}}
\newcommand{\pragsourceonly}{\texttt{Source-Rec-Only}}

\newcommand{\optimalpragans}{$\texttt{Answer-Rec}^*$}
\newcommand{\optimalpragsource}{$\texttt{Source-Rec}^*$}
\newcommand{\optimalpragas}{$\texttt{Ans-Src-Rec}^*$}

\newcommand{\qmsum}{QMSum\ }
\newcommand{\multioped}{MultiOpEd\ }
\newcommand{\squality}{SQuALITY\ }

\title{Does This Summary Answer My Question? \\ Modeling Query-Focused Summary Readers with Rational Speech Acts}

\author{
Cesare Spinoso-Di Piano \bf and Jackie Chi Kit Cheung \\
Mila – Quebec Artificial Intelligence Institute \\
McGill University \\
\texttt{cesare.spinoso-dipiano@mail.mcgill.ca}
}

\begin{document}

\maketitle

\begin{abstract}
    Query-focused summarization (QFS) is the task of generating a summary in response to a user-written query. Despite its user-oriented nature, there has been limited work in QFS in explicitly considering a user's understanding of a generated summary, potentially causing QFS systems to underperform at inference time. In this paper, we adapt the Rational Speech Act (RSA) framework, a model of human communication, to explicitly model a reader's understanding of a query-focused summary and integrate it within the generation method of existing QFS systems. In particular, we introduce the \emph{answer reconstruction} objective which approximates a reader's understanding of a summary by their ability to use it to reconstruct the answer to their initial query. Using this objective, we are able to re-rank candidate summaries generated by existing QFS systems and select summaries that better align with their corresponding query and reference summary. More generally, our study suggests that a simple and effective way of improving a language generation system designed for a user-centered task may be to explicitly incorporate its user requirements into the system's generation procedure.
\end{abstract}

\section{Introduction}

\begin{figure}[t]
    \centering
    \includegraphics[width=\linewidth]{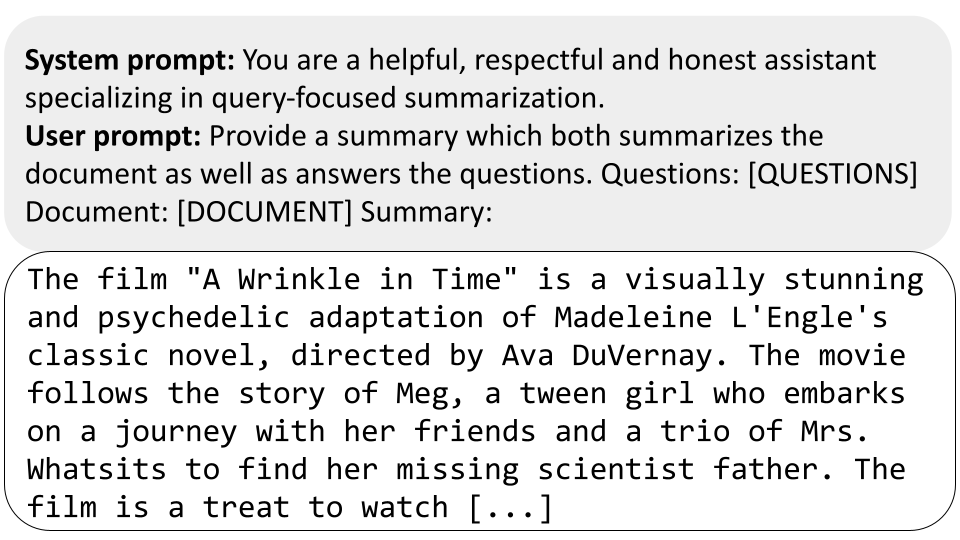}
    \caption{An example of a \emph{query-focused summary} (truncated) generated by Llama 3 for this \href{https://www.thewrap.com/a-wrinkle-in-time-film-review-oprah-winfrey-ava-duvernay-2018/}{\underline{article}} on the movie ``A Wrinkle in Time'' from the \multioped dataset. The only question which was asked about this article ([QUESTIONS]) was ``Is ``A Wrinkle in Time'' worth watching?''. The full prompt we give to Llama 3 can be found in Appendix~\ref{sec:llm_prompts}.}
    \label{fig:example-qfs-llama-3}
\end{figure}

In automatic summarization, there has been a growing interest in pushing summarization towards more user-centered objectives \citep{Daz2007UsermodelBP,Lerman2009SentimentSE,Yan2011SummarizeWY,Bhatnagar2023IWT}. One popular approach has been to provide summarization systems with additional user preferences, such as user queries \citep{Dang2005OverviewOD,daumeBayesianQueryfocusedSummarization2006,Baumel2018QueryFA,vigExploringNeuralModels2022}. In this redefined query-focused summarization (QFS) setting, a system should provide a concise and factual summary of a source document while additionally providing the necessary information for the user to answer their initial query (or queries). For instance, in Figure~\ref{fig:example-qfs-llama-3}, a Llama 3 model is tasked with summarizing an article about the movie ``A Wrinkle in Time'' while answering the question ``Is ``A Wrinkle in Time'' worth watching?''.

Despite the more user-oriented nature of QFS, most QFS systems limit their considerations regarding how their generated summaries may be perceived and used by their readers. In particular, during the generation process, most QFS systems do not \emph{explicitly} consider the extent to which their generated summary will allow the user to actually answer their initial query. Instead, existing QFS systems and their generation procedures consider their users indirectly by, for instance, integrating query-answer relevance scores into their system's fine-tuning stage \citep{Sotudeh2023QontSumOC}. Meanwhile, work in personalized summarization has already shown that explicitly considering ``user characteristics'' leads to summaries which better account for user needs \citep{Li2019TowardsPR}.


As a result, in this paper, we seek to provide an explicit model of the reader of a query-focused summary. We hypothesize that this explicit user consideration will enable existing QFS systems to generate summaries which better align with how they will be read, understood and used. To do so, we take the approach of modeling how the user might understand the generated summary in the context of the posed query. In particular, given a source document and one or more related queries, we consider a good summary to be one that allows the user to provide a high-quality answer to the query without needing to consult the source document. To model this, we introduce the \emph{answer reconstruction} score which is the likelihood of generating the answer to the initial query conditioned on the summary \emph{and not} the source document. This answer reconstruction score is then used at inference time to re-rank a pool of candidate summaries generated by a QFS system.


This model of the reader of the summary is inspired by recent work which has attempted to leverage the Rational Speech Act (RSA) theory \citep{degen_rational_2023} to model pragmatic language phenomena such as conversational implicature. In RSA, a speaker attempts to convey a piece of information to a listener by generating an utterance which the speaker believes will be easily and unambiguously understood by the listener. To do so, the speaker, often referred to as the \emph{pragmatic speaker}, will generate an utterance whose intended meaning is likely to be recovered by the listener, often referred to as the \emph{pragmatic listener}. Translating this to the context of QFS and the answer reconstruction objective, a \emph{pragmatic summarizer} will produce a summary which can be directly used by a \emph{pragmatic reader} to answer their initial query. 


As a result of our adaptation of the RSA framework, we show that standard decoding methods for generating summaries with QFS systems underperform against generation procedures which explicitly consider the readers of the summaries. More specifically, we test our reader model with an end-to-end fine-tuned summarization model and a prompt-based summarization model and show that we are able to select summaries which better match their corresponding reference summaries in terms of ROUGE and other standard metrics. Beyond QFS, our results suggest that current language generation systems designed to account for user preferences may be missing out on potential performance gains by opting not to explicitly consider users during the text generation stage.


To summarize, in this paper, the research question we answer is: ``Does an explicit model of a reader produce query-focused summaries which better align with their corresponding query and reference summary?''. To this end, we extend existing work in query-focused summarization (QFS) by adapting the Rational Speech Act (RSA) framework to provide an explicit model of the reader of a query-focused summary. In particular, we develop an \emph{answer reconstruction} objective which measures and re-ranks candidate summaries based on their suitability in providing the necessary information to answer the initial query. We show that the additional answer reconstruction step of our generation method leads to selecting candidate summaries which are more appropriate. More generally, our work provides the community interested in building user-centered language generation systems with evidence that explicitly integrating user models into the generation process may lead to systems which better meet user needs.

\section{Related Work}

\paragraph{Query-focused summarization (QFS)} Some of the first concerted efforts in developing QFS systems date back to \citet{Dang2005OverviewOD,Dang2006OverviewOD} where QFS was presented as a shared task. Since then, QFS has received increasing attention due, in part, to it being more user-centered in nature than traditional generic summarization \citep{kryscinski2019neural,vigExploringNeuralModels2022}.

There have been several proposed modeling approaches to QFS, the two most common approaches being a two-step retrieval-abstraction method and an end-to-end method \citep{vigExploringNeuralModels2022}. In retrieval-abstraction methods such as \citet{Egonmwan2019CrossTaskKT}, the summarization system first uses the provided query to retrieve the most relevant sentences from the document to be summarized and subsequently passes those sentences to an abstractive module tasked with providing a concise and factual overview of those sentences. In end-to-end methods \citep{Baumel2018QueryFA,Xie2020ConditionalSF}, a summarization system is fine-tuned on source document, question and reference summary triples allowing it to directly generate a summary from a new source document and question pair at inference time. Other notable approaches which do not cleanly fit within these two paradigms include \citet{Xu2020GeneratingQF} which leverage generic summaries to improve QFS systems and \citet{Xu2021TextSW} which develop a latent query model.

In closer connection with our work, there have been several methods which explicitly consider the answers to queries in their QFS systems. In \citet{Su2021ImproveQF}, the authors compute scores for each sentence in a source document based on their relevance to the answer of a query and integrate those scores within the attention computation of a BART-based end-to-end summarization system. In \citet{Xu2020CoarsetoFineQF}, the authors use several scoring modules to iteratively refine the relevance of source document sentences as answers to the query before passing them to an abstractive summarization module. Another approach by \citet{Zhao2023QTSummQS} has been to integrate the answer produced by a QA system as one of the inputs to an end-to-end summarization system. In contrast, we propose to score candidate summaries based on a modeled reader's ability to answer the initial question based on the summary alone. Thus, our method may be used with existing QFS systems by generating several candidate summaries and selecting the most appropriate summary at inference time.

\paragraph{Rational speech act (RSA)} Our model of the readers of a summary is inspired by the RSA framework, which has been used to model several language generation tasks to better reflect human communication\citep{grice1975logic,Frank2012PredictingPR,goodman2013knowledge,Bergen2016PragmaticRT}. In \citet{andreasReasoningPragmaticsNeural2016}, the authors model the task of informative image captioning by using, for the first time, \emph{neural} pragmatic listeners and speakers. \citet{friedUnifiedPragmaticModels2018} extend the use of RSA to instruction following and generation. \citet{shen-etal-2019-pragmatically} investigate RSA for generic summarization and language generation from structured meaning representations while \citet{darrin-etal-2024-glimpse} leverage the RSA framework to develop a multi-document summarization system of scholarly reviews.

We close this section by contrasting our approach with a similar adaption of RSA to summarization. In \citet{shen-etal-2019-pragmatically}, the authors introduce the \emph{source reconstruction} objective whereby the goodness of a summary is measured by a reader's ability to reconstruct the source document given the generated summary. Convenient as it may be, we do not believe source reconstruction to be a realistic model of a summary's reader. For instance, a summarization system may correctly choose to drop unimportant information from the source text in the summary making it impossible to perfectly reconstruct the source. In contrast, our method's answer reconstruction objective is directly defined through a query-focused summary's intended use case: as a way of informing the user about their posed query.

\section{Answer Reconstruction: An RSA Model of a QFS Reader}

\label{sec:answer_reconstruction}

We formalize our explicit model of a query-focused summary reader firstly by discussing the RSA framework in general and subsequently by formally discussing how we adapt it to model a query-focused summary reader through the answer reconstruction objective. The intuition and driving principle behind our proposed method is that user-oriented language generation tasks such as QFS involve user requirements which can naturally be integrated into the generation procedure through the RSA framework. We hypothesize that this explicit consideration of users will favour outputs which better align with their expectations of user-centered language generation systems.

\paragraph{RSA and human communication} The RSA framework was originally introduced to model human communication and, more specifically, to explicitly account for a listener in the context of modeling pragmatic phenomena like conversational implicature \citep{Frank2012PredictingPR,goodman2013knowledge,Bergen2016PragmaticRT}. In this framework, a pragmatic speaker, $S_1$, attempts to convey a message with some meaning, $m$, by generating an utterance, $u$. To do so, the pragmatic speaker uses a \emph{literal speaker}, $S_0$, to first generate candidate utterances $\mathcal{U}$ where the utterances $u \in \mathcal{U}$ are generated from a uniform distribution, $P_{S_0}(U | M = m)$, over utterances with literal meaning matching $m$. The pragmatic speaker then explicitly posits a pragmatic listener, $L_1$, to approximate the usefulness of each candidate utterance $u \in \mathcal{U}$. In particular, $S_1$ re-scores each candidate utterance, $u \in \mathcal{U}$, by using the probability that $L_1$ recovers the intended meaning of their message when observing $u$, $P_{L_1}(M = m | U = u)$. Finally, the pragmatic speaker combines these two probabilities with a rationality parameter $\lambda$ to provide a final score for each utterance $u \in \mathcal{U}$ represented by\footnote{We drop the $P_{S_1}$ notation for the pragmatic speaker as, except for $\lambda = 0,1$, this product does not necessarily induce a valid probability distribution.}
\begin{equation}
    S_1(u|m;\lambda) = P_{S_0}(u|m)^{1 - \lambda} \cdot P_{L_1}(m|u)^\lambda
\end{equation}
\paragraph{RSA and QFS} To adapt the RSA model to QFS, we provide new definitions for the speakers, $S_0,S_1$, and listener, $L_1$, which we refer to as summarizers, $S_0,S_1$, and reader, $R_1$, respectively. In addition, we base the intended meaning of a query-focused summary on its suitability in providing enough information for $R_1$ to correctly answer their initial query.

In our adapted RSA framework, we define the literal summarizer as any QFS system which takes as input some source text $x$ and query $q$ and which induces a probability distribution over candidate summaries, $P_{S_0}(Y|X = x, Q = q)$\footnote{We leave the consideration of multiple source documents as well as multiple user queries to future work.}. We think of existing QFS systems as literal summarizers since, similar to the original RSA formulation, they do not directly consider the readers of their summaries. In our formulation, given $x$ and $q$, $S_1$ leverages $S_0$ as a way to sample $n$ candidate summaries:
\begin{equation}
    \hat{y_i} \sim P_{S_0}(Y | X = x, Q = q), i = 1 \dots n
\end{equation}
To implement our definition of the intended meaning of a query-focused summary, we introduce the \emph{answer reconstruction} objective. This objective allows re-scoring a candidate summary $\hat{y_i}$ based on $R_1$'s ability to generate the answer to the original query $q$ by using $\hat{y_i}$ instead of $x$. To do so, we suppose that $S_1$ has access to a question-answer (QA) generation system $F$ which takes as input a document $d$ and a related question $q^*$\footnote{We use $d$ and $q^*$ in our QA notation to disambiguate with our QFS notation for the source text $x$ and its query $q$.} and provides an answer $F(D = d, Q^* = q^*) = a$ along with the probability for that answer, $P_F(A = a | D = d, Q^* = q^*)$. 

Thus, given $x$ and $q$, $S_1$ models $R_1$'s understanding of $\hat{y_i}$ by generating an answer $a = F(x, q)$ and by evaluating the probability of $R_1$ reconstructing $a$ via the same QA process. That is, 
\begin{multline}
\label{eq:answer_reconstruction}
    P_{R_1}(A = a | \hat{Y} = \hat{y_i}, Q = q) \coloneq \\ P_F(A = a  | D = \hat{y_i}, Q^* = q)
\end{multline}
Finally, a rationality parameter $\lambda \in [0,1]$ is used by $S_1$ to regulate the importance assigned to $R_1$'s answer reconstruction ability in selecting the final summary. That is, $S_1$'s final score of a candidate summary $\hat{y_i}$ for a source text $x$ with query $q$ and corresponding answer $a = F(x, q)$ will be:
\begin{equation}
    S_1(\hat{y_i}|x,q;\lambda) = P_{S_0}(\hat{y_i}|x, q)^{1 - \lambda} \cdot P_{R_1}(a|\hat{y_i},q)^\lambda,
\end{equation}
where we notice that when $\lambda = 1$ a summary's score is entirely determined by $R_1$'s ability to answer $q$ while when $\lambda = 0$ the summary's score is the one provided by the original QFS system. Thus, we hypothesize that an explicit consideration of $R_1$, i.e. $\lambda > 0$, rather than an implicit consideration, i.e. $\lambda = 0$, may lead to the selection of candidate summaries which better match their reference summaries.

\section{Experiments}

\subsection{Datasets}

\label{sec:datasets}

We describe the datasets that we use to run our experiments on our QFS setting. Notably, we do not use the DUC datasets \citep{dangDUC2005Evaluation2006} since we limit our focus to single-document and single-query-focused summarization. We provide summary statistics for each dataset in Table~\ref{tab:summ_stats}.

\paragraph{\multioped} The \multioped dataset \citep{liu-etal-2021-multioped} is a collection of news editorials containing opinion articles and summaries for each article which also provide the answer to a specific question.

\smalltodo{ \multioped has gold answers. Consider incorporating this as a baseline in a next revision.}

\paragraph{\qmsum} The \qmsum dataset \citep{qmsum} is a query-focused meeting summarization dataset where the source texts are excerpts from different meeting transcripts. Each excerpt has an associated question and reference summary summarizing the excerpt while providing an answer to the question.

\paragraph{\squality} The \squality dataset \citep{Wang2022SQuALITYBA} is a collection of short stories with multiple queries and corresponding query-focused summaries. 

\begin{table}[H]
    \centering
    \resizebox{\linewidth}{!}{
\begin{tabular}{lcccc}
    \toprule
     \multirow{3}{*}{\begin{tabular}{@{}l@{}}Dataset \\ Name\end{tabular}} & \multirow{3}{*}{\begin{tabular}{@{}l@{}}Test \\ Size\end{tabular}} & \multicolumn{3}{c}{Average length} \\
    & & Source & Question & \begin{tabular}{@{}c@{}} Reference \\ Summary \end{tabular} \\
    \cmidrule{3-5}
    \multioped & 531 & 925 & 7 & 102 \\
    \qmsum & 1365 & 893 & 13 & 63 \\
    \squality & 252 & 4986 & 9 & 228 \\ 
    \bottomrule
\end{tabular}
    }
    \caption{Summary statistics for each of the datasets. Test size refers to the number of unique source-question pairs in the test set of each dataset and the average length refers to the average space-separated string lengths of different dataset fields.}
    \label{tab:summ_stats}
\end{table}


\subsection{Answer-Reconstruction-Based QFS Pipeline}

To test our answer-reconstruction-based model of a QFS reader, we implement a pragmatic summarizer, \pragans, which leverages a literal summarizer to generate candidate summaries and a QA system to re-rank them based on their suitability as answers to the initial query. In particular, for each dataset and source document-question pair, we leverage a literal summarizer to generate a set of 10 candidate summaries. In parallel, we leverage an existing QA generation system to generate answers for each source document and question pair. We then combine the likelihood scores produced by the literal summarizer, $P_{S_0}$, and the answer reconstruction score, $P_{R_1}$, as defined in Equation~\ref{eq:answer_reconstruction} to re-rank the set of candidate summaries. As a result, the summary selected by our pragmatic summarizer is:
\begin{equation}    
    \argmax_{i=1 \dots 10} \underbrace{P_{S_0}(\hat{y_i}|x,q)^{1 - \lambda} \cdot P_{R_1}(a|\hat{y_i}, q)^\lambda}_{S_1(\hat{y_i}|x,q;\lambda)}
\end{equation}
We use multiple values of $\lambda$ which we further detail in Appendix~\ref{sec:hyperparameters}.

We provide additional details related to our summarization and QA systems in the next subsections. All the model checkpoints we use for our summarization and QA systems are taken from HuggingFace\footnote{\url{https://huggingface.co/}} and all fine-tuning is done through PyTorch Lightning.\footnote{\url{https://www.pytorchlightning.ai}} We run fine-tuning and inference on a variable number of NVIDIA Quadro RTX 8000 48GB GPUs based on the GPU memory requirements of the corresponding model architecture and setting.\footnote{Code can be found here: \url{https://github.com/cesare-spinoso/RSASumm-CustomNLP4U}.}

\subsubsection{Literal Summarizers}

We use BART and Llama 3 as the literal summarizers for all of our datasets. We experiment with several standard decoding methods including standard sampling, top-k sampling \citep{Fan2018HierarchicalNS} and nucleus sampling \citep{Holtzman2019TheCC}. Since summary selection is only meaningful when selecting from a \emph{diverse} pool of candidate summaries, we follow decisions by \citet{Holtzman2019TheCC} and \citet{Aralikatte2021FocusAP} to generate diverse summaries by setting $k = 640$ in top-k sampling and $p = 0.95$ in nucleus sampling. Further decoding hyperparameters are discussed in Appendix~\ref{sec:hyperparameters}. In addition, we use the conditional likelihood as computed by each model to implement $P_{S_0}$.\tinytodo{Add some kind of beam search in next revision.}

\paragraph{BART} BART \citep{bart_model} is an encoder-decoder Transformer model trained with a denoising objective function and is commonly used for language generation tasks including QFS \citep{Su2021ImproveQF,vigExploringNeuralModels2022}. We use the \texttt{facebook/bart-large-cnn} checkpoint of BART which is fine-tuned on the CNN/Daily Mail summarization dataset and further fine-tune it on the training/validation split of each of our tested datasets. We follow existing QFS conventions \citep{vigExploringNeuralModels2022} and concatenate the user query and the source text with the model's delimiter token. We truncate the input from the right so that it fits within BART's 1024 maximum input token limit. Additional hyperparameter configurations are discussed in Appendix~\ref{sec:hyperparameters}.

\paragraph{Llama 3} Llama 3 is a LLM developed by Meta \citep{Dubey2024TheL3}. We use the \texttt{meta-llama/Meta-Llama-3-8B-Instruct} checkpoint which is the 8 billion parameter version fine-tuned with both instruction tuning and reinforcement learning from human feedback (RLHF). This class of models has been shown to be competitive on summarization benchmarks \citep{yangExploringLimitsChatGPT2023,zhangBenchmarkingLargeLanguage2024}. We use 8-bit quantization for inference and truncate the source document from the right so that it fits within Llama 3's context window of 4096 tokens. The prompt we use for summary generation can be found in Appendix~\ref{sec:llm_prompts}.

\subsubsection{Question-Answer Generation}

Given the reported versatility of Llama 3 \citep{Dubey2024TheL3}, we rely on the same checkpoint as the literal summarizer to implement the QA system $F$ and its probability distribution $P_F$. To generate an answer to a question $q$ about a source document $x$, $a = F(q, x)$, we prompt Llama 3 with a QA prompt (Appendix~\ref{sec:llm_prompts}) and use beam search decoding with a beam size of 5. The pragmatic reader's answer reconstruction score for a candidate summary $\hat{y}$ is thus the probability of Llama 3 generating $a$ given the same prompt, but with $x$ replaced with $\hat{y}$.

We provide examples of generated answers for each dataset in Table~\ref{tab:summ_examples}.

\begin{table}[H]
\centering
\resizebox{\linewidth}{!}{
\begin{tabular}{lp{4cm}p{4cm}}
\toprule
Dataset Name & Question & Generated Answer \\
\midrule
\multioped & Is ``A Wrinkle in Time'' worth watching? & Yes, ``A Wrinkle in Time'' is worth watching. \\
\qmsum & What did PhD F think about computational resources? & PhD F thought that soon, when they get all the new machines up, they will have lots more compute to use. \\
\squality & Who is murra foray and how is she significant to the story? & Murra Foray is the First Counselor of the Travelers Aid Bureau. \\
\bottomrule
\end{tabular}
}
\caption{Examples of questions and answers, as generated by Llama 3, for each dataset we use.}
\label{tab:summ_examples}
\end{table}

\subsection{Evaluation}

To measure summarization quality, we evaluate our query-focused summarization pipeline using ROUGE-1, ROUGE-2, ROUGE-L \citep{lin-2004-rouge}, METEOR \citep{banarjee2005}, and BERTScore \citep{zhang-bertscore} scores between the summary selected by the pragmatic summarizer and the reference summary. In the case that several reference summaries have been written for the same source text-question pair, we take the maximum metric scores across the generated-reference summary pairs.

In addition, to measure text quality, we leverage the learnt reference-free SEAHORSE metrics developed by \citet{clark-seahorse}. These metrics are trained on source text, summary and human judgment triples and include 6 evaluation categories, 4 of which measure text quality. \textbf{Comprehensible} measures summary fluency. \textbf{Repetition} checks for lack of repetition in a summary. \textbf{Grammar} measures summary grammaticality. \textbf{Conciseness} measures summary succinctness.\smalltodo{Next version should have human evals. Design with Jackie.}

\subsection{Alternative Pragmatic Summarizers}

We implement alternative ways for $S_1$ to select the final summary as a way of comparing the effectiveness of our answer reconstruction model of a QFS reader.\smalltodo{Q: Should the title of the different S1's be capitalized?}

\paragraph{Oracle} For each pool of 10 candidate summaries, the $\pragoracle$ pragmatic summarizer selects the summary which maximizes the metric being used to evaluate the system.

\paragraph{Random} The $\pragrandom$ pragmatic summarizer randomly selects one of the 10 candidate summaries as the final summary.

\paragraph{Source reconstruction} The source reconstruction pragmatic summarizer, $\pragsource$, as proposed by \citet{shen-etal-2019-pragmatically} uses an alternative reader model which scores a candidate summary $\hat{y}$ based on the conditional likelihood of reconstructing the source document $x$. We use Llama 3 and a source reconstruction prompt (Appendix~\ref{sec:llm_prompts}) and denote this likelihood as $P_{R_1}(x|\hat{y})$ where we thus have that
\begin{equation}
    S_1(\hat{y}|x,q;\lambda) = P_{S_0}(\hat{y}|x,q)^{1 - \lambda} \cdot P_{R_1}(x|\hat{y})^\lambda
\end{equation}
\paragraph{Answer-source reconstruction} We believe that the answer and source reconstruction objective may be somewhat complementary in modeling the QFS reader. To verify this hypothesis, we also experiment with an answer-source reconstruction, $\pragas$, which introduces an additional interpolating parameter $\alpha \in [0, 1]$ for the answer reconstruction likelihood, $P_{R_1}(a|\hat{y},q)$, and the source reconstruction likelihood, $P_{R_1}(x|\hat{y})$. Thus, we have that
\begin{multline}    
    S_1(\hat{y_i}|x,q;\lambda,\alpha) = P_{S_0}(\hat{y_i}|x,q)^{1 - \lambda} \\ \cdot \left ( P_{R_1}(a|q,\hat{y_i})^{1 - \alpha} \cdot P_{R_1}(x|\hat{y_i})^\alpha \right )^\lambda
\end{multline}
\section{Results}

\subsection{Summary Quality}

We present the ROUGE, METEOR and BERTScore results of the various pragmatic summarizers. In particular, we present the performance of pragmatic summarizers which select summaries solely based on the literal summarizer's likelihood scores ($\literalsumm$) or the pragmatic reader's reconstruction score ($\pragansonly$ and $\pragsource$). We also provide the performance of the optimal pragmatic summarizers (\optimalpragans, \optimalpragsource\, and \optimalpragas) by selecting optimal values of $\lambda$ and $\alpha$. We show the performance of pragmatic summarizers on the \multioped dataset in Table~\ref{tab:multioped_nucleus_0.95_temp_1.2}.  The summaries generated by $S_0$ in this table were produced using nucleus sampling with $p = 0.95$ and a temperature of $1.2$. Results for our other tested datasets and decoding techniques can be found in Appendix~\ref{sec:app_add_results_alternative_prags}.

\begin{table*}[h]
\centering
\resizebox{\linewidth}{!}{
\begin{tabular}{lcccccccccc}
\toprule
& \multicolumn{5}{c}{BART} & \multicolumn{5}{c}{Llama 3} \\
 & R-1 & R-2 & R-L & METEOR & BERTScore & R-1 & R-2 & R-L & METEOR & BERTScore \\
\cmidrule(r){2-6}\cmidrule(r){7-11}

 $\pragrandom$                    & 0.3378            & 0.0689            & 0.1777            & 0.2547               & 0.1904                  & 0.3120               & 0.0683               & 0.1674               & 0.2751                  & 0.1734                     \\\cmidrule(r){2-6}\cmidrule(r){7-11}
 $\literalsumm$            & 0.3396            & \textbf{0.0712}   & \textbf{0.1810}   & 0.2464               & \textbf{0.2040}         & 0.2915               & 0.0666               & 0.1583               & 0.2823                  & 0.1558                     \\
 $\pragansonly$             & 0.3382            & 0.0683            & 0.1769            & 0.2544               & 0.1825                  & 0.3092               & 0.0683               & 0.1659               & 0.2746                  & 0.1678                     \\
 $\pragsourceonly$          & 0.3347            & 0.0670            & 0.1745            & 0.2536               & 0.1795                  & 0.3087               & 0.0688               & 0.1658               & 0.2752                  & 0.1718                     \\\cmidrule(r){2-6}\cmidrule(r){7-11}
 \optimalpragans             & \textbf{0.3400}   & 0.0709            & 0.1793            & \textbf{0.2622}      & 0.1962                  & 0.3316               & 0.0708               & \textbf{0.1771}      & 0.2801                  & 0.1975                     \\
 \optimalpragsource         & 0.3396            & 0.0711            & 0.1810            & 0.2539               & 0.2038                  & 0.3118               & 0.0693               & 0.1674               & \textbf{0.2826}         & 0.1793                     \\
 \optimalpragas & 0.3386            & 0.0705            & 0.1799            & 0.2616               & 0.2022                  & \textbf{0.3320}      & \textbf{0.0710}      & 0.1771               & 0.2823                  & \textbf{0.1978}            \\\cmidrule(r){2-6}\cmidrule(r){7-11}
 $\pragoracle$                    & 0.3913            & 0.1030            & 0.2117            & 0.3079               & 0.2496                  & 0.3568               & 0.0953               & 0.1966               & 0.3150                  & 0.2285                     \\
\bottomrule
\end{tabular}
}
\caption{Performance of the different pragmatic summarizers on the \multioped dataset measured using the ROUGE, METEOR, and, BERTScore metrics. The summaries generated by the literal summarizer used nucleus sampling ($p = 0.95$, temperature of 1.2).}
\label{tab:multioped_nucleus_0.95_temp_1.2}
\end{table*}

\begin{table}[h]
\centering
\resizebox{\linewidth}{!}{
\begin{tabular}{lccccc}
\toprule
                                  &   R-1 &   R-2 &   R-L &   METEOR &   BERTScore \\
\cmidrule(r){2-6}
 $\pragrandom$                    &     0 &     1 &     0 &        0 &           0 \\\cmidrule(r){2-6}
 $\literalsumm$            &     2 &     2 &     4 &        0 &           2 \\\cmidrule(r){2-6}
 $\pragansonly$             &     0 &     0 &     0 &        0 &           0 \\
 $\pragsourceonly$          &     0 &     0 &     0 &        0 &           0 \\
 \optimalpragans             &     2 &     1 &     2 &        5 &           2 \\
 \optimalpragsource         &     2 &     3 &     1 &        3 &           3 \\
 \optimalpragas &     4 &     3 &     3 &        2 &           3 \\
\bottomrule
\end{tabular}
}
\caption{Number of times each pragmatic summarizer achieves the highest ROUGE, METEOR, and BERTScore scores on the \multioped dataset. We aggregate counts across all combinations of models (Llama 3 and BART) and decoding types (Standard, nucleus, and top-k sampling).}
\label{tab:multioped_prag_summ_top_score_freq}
\end{table}

Across all datasets, models and decoding techniques, we observe that reader-unaware pragmatic summarizers which only leverage the scores of literal summarizers consistently underperform against reader-aware pragmatic summarizers. This finding is more clearly illustrated through Table~\ref{tab:multioped_prag_summ_top_score_freq} which aggregates the \multioped results from Table~\ref{tab:multioped_nucleus_0.95_temp_1.5} and Appendix~\ref{sec:app_add_results_alternative_prags} by counting, for each model and decoding technique, the number of times each pragmatic summarizer variant achieves the highest metric score. Similar tables for \qmsum and \squality are shown in Appendix~\ref{sec:app_prag_summ_top_score_freq}.

To better understand these results, we present the performance changes of the different variants of pragmatic summarizers relative to $\pragrandom$ on the \multioped dataset in Table~\ref{tab:multioped_relative_to_random_scores_short}. More detailed tables for all the datasets can be found in Appendix~\ref{sec:app_relative_change_scores}. We observe that, in the majority of cases, using a pragmatic summarizer which scores summaries solely based on the answer objective (i.e., \pragansonly) leads to worse than random performance. This is unsurprising given that we expect query-focused reference summaries to contain information beyond the answer to the initial query. Surprisingly, however, we also observe worse-than-random performance in reader-unaware summarizers (i.e., \literalsumm). This finding suggests that the likelihood scores of QFS summaries do not faithfully reflect the goodness of their generated summaries.
\begin{table}[h]
\centering
\resizebox{1\linewidth}{!}{
\begin{tabular}{lcccccc}
\toprule
& \multicolumn{3}{c}{BART} & \multicolumn{3}{c}{Llama 3} \\
 & R-1 & METEOR & BERTScore & R-1 & METEOR & BERTScore \\
\cmidrule(r){2-4}\cmidrule(r){5-7}

 $\literalsumm$            & $+$0.20             & $-$\underline{0.69}  & $+$1.66                 & $-$\underline{1.14}  & $+$0.93                 & $-$\underline{1.88}        \\\cmidrule(r){2-4}\cmidrule(r){5-7}
 $\pragansonly$             & $-$0.29             & $-$0.54              & $-$0.85                 & $-$0.25              & $-$\underline{0.09}     & $-$1.16                    \\
 $\pragsourceonly$          & $-$\underline{0.31} & $-$0.44              & $-$\underline{1.02}     & $-$0.44              & $+$0.03                 & $-$0.46                    \\\cmidrule(r){2-4}\cmidrule(r){5-7}
 \optimalpragans             & $-$0.08             & $+$\textbf{0.06}     & $+$0.66                 & $+$1.02              & $+$0.55                 & $+$\textbf{2.31}           \\
 \optimalpragsource         & $+$\textbf{0.21}    & $-$0.39              & $+$\textbf{1.70}        & $-$0.26              & $+$\textbf{0.94}        & $+$0.37                    \\
 \optimalpragas & $+$0.17             & $+$0.01              & $+$1.54                 & $+$\textbf{1.04}     & $+$0.92                 & $+$2.30                    \\
\bottomrule
\end{tabular}
}
\caption{Relative change in ROUGE-1, METEOR, and BERTScore scores between the pragmatic summarizers and $\pragrandom$ on the \multioped dataset. Relative changes are aggregated across all decoding methods. A value of $+x$ ($-x$) indicates that the average metric score achieved is $x\%$ higher (lower) than the one achieved by $\pragrandom$. The lowest relative change is \underline{underlined}, and the highest is \textbf{bolded}.}
\label{tab:multioped_relative_to_random_scores_short}
\end{table}

Moreover, we observe that using a rationality parameter which optimally interpolates between $S_0$ and $R_1$ consistently leads to performance improvements as compared to the literal summarizer. In particular, when optimizing for $\lambda$ and $\alpha$, \optimalpragas\, achieves the highest metric scores across all datasets, literal summarizers, and decoding techniques in $39\%$ of cases, \optimalpragans\, in $26\%$ of cases and \pragsource\, in $21\%$ of cases. In addition, these three pragmatic summarizers surpass random selection in $87\%$, $80\%$, and $60\%$ of cases, respectively. In contrast, the reader-unaware pragmatic summarizer outperforms random selection in only $40\%$ of cases.

Overall, these results suggest that an explicit model of query-focused summary readers benefits the QFS generation process, but that the extent of these benefits heavily relies upon how the reader is modeled. We see that the answer reconstruction objective, which reflects the downstream use of query-focused summaries, is a more beneficial reader model than the generic source reconstruction objective. Moreover, we find that these reconstruction objectives are somewhat complementary and that combining them may lead to the best results. This might be because the answer reconstruction objective favours summaries which explicitly address the initially posed query, while the source reconstruction objective prevents the selection from degenerating to a simple QA-like response.

\subsection{Text Quality}

We evaluate the text quality of selected summaries using the SEAHORSE metrics. In particular, we are interested in measuring the tradeoff between summarization quality and text quality which has been reported in other work which has applied RSA to text generation \citep{andreasReasoningPragmaticsNeural2016,darrin-etal-2024-glimpse}. We show this tradeoff by plotting the different values of the summarization and text quality metrics across values of $\lambda$ for both $\pragans$ and $\pragsource$ in Figure~\ref{llama3_qfs_temp_2_multioped_metrics_versus_lambda}. Additional tables and plots can be found in Appendix~\ref{sec:app_text_quality}.

\begin{figure}[h]
            \centering
            \includegraphics[width=\linewidth]{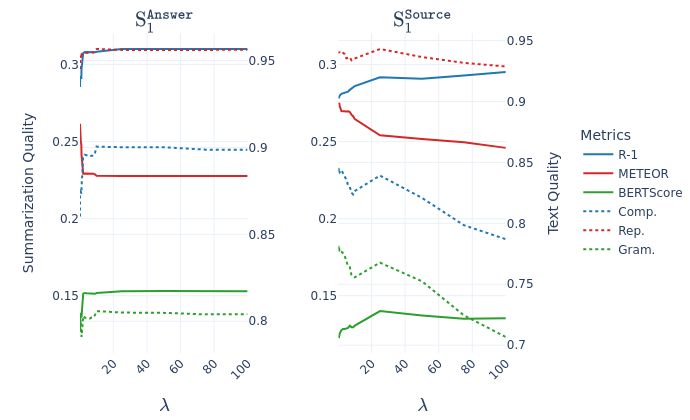}
            \caption{Tradeoff between summarization quality and text quality as controlled by $\lambda$ for the \multioped dataset using Llama 3 with standard sampling (temperature of 2). Solid lines represent summarization quality metrics (ROUGE-1, METEOR, and BERTScore), while dashed lines represent text quality metrics (Comprehensibile, Repetition, and Grammar). Note that the left and right y-axes have different scales for summarization quality and text quality, respectively. }
            \label{llama3_qfs_temp_2_multioped_metrics_versus_lambda}
        \end{figure}

Interestingly, while there is a drop in text quality as $\lambda$ is increased for $\pragsource$, this drop does not occur for $\pragans$ even when studying the literal summarizers which use the highest sampling temperature. Intuitively, this may be because summaries which optimize for source reconstruction will cover more information which naturally increases the complexity of sentences and, thus, the likelihood for them to contain grammatical errors. On the other hand, the more to-the-point nature of the answer reconstruction objective may favour more concise summaries.



\section{Conclusion}

In conclusion, we have argued that explicitly modeling readers of query-focused summaries can produce summaries which better align with user requirements. Specifically, we demonstrated that modeling QFS readers using the RSA framework with a task-specific answer reconstruction objective leads to summaries which align more closely with reference summaries compared to those selected with either a generic source reconstruction objective or the likelihood scores of existing QFS systems. In addition, we have shown that existing reader-unaware QFS systems may not be fully realizing their potential to generate summaries which best reflect their users' needs. Augmenting their generation procedure with a hybrid answer-source reconstruction objective as a model of the reader may be a way to close this gap in performance.  Beyond QFS, our findings suggest that existing language generation systems could better serve their users by explicitly considering them during the language generation process rather than implicitly through the patterns learnt during the finetuning stage.

\section*{Limitations}

In this work, we aimed to showcase the benefit of modeling query-focused summary readers by augmenting existing QFS systems with a reader-aware generation process. Due to resource constraints, we only considered two end-to-end QFS systems, BART and Llama 3, and no two-step retrieval-abstraction QFS systems. We leave the exploration of other existing QFS systems as future work. Moreover, we note that noise is inevitably introduced by using a QA generation system rather than gold answers in our reader model. However, we believe that this is a reasonable limitation given that it is unrealistic to assume access to gold answers at inference time.

In addition, though they are commonly used throughout the summarization and language generation community, we acknowledge that our evaluation methods suffer from well-documented validity issues. For instance, the goodness of a generated summary and its suitability for a reader may not correlate with word overlap metrics (ROUGE, METEOR) between reference and generated summaries. In addition, the SEAHORSE learnt metrics we use are working in an out-of-distribution manner since they were originally designed for multifaceted summaries. We plan on addressing these concerns in future work by running an extensive human evaluation study on QFS systems. This study will allow us to compute metrics such as the correlation between the ranks of generated summaries according to our different pragmatic summarizers and the one proposed by humans.

\section*{Ethical Considerations}

Our method does not address the inherent risks of deploying QFS systems in real-world scenarios, such as in informing users on sensitive matters or in high-stakes decision-making situations. Moreover, we acknowledge the risk posed by abstractive generation systems which are known to be prone to hallucinations. In general, we think of QFS systems as tools to help users quickly iterate over large collections of documents, with safeguards in place to allow users to verify their factuality such as attribution methods.

\bibliography{custom}

\clearpage

\appendix

\section{Inference and Fine-Tuning Hyperparameters}

\label{sec:hyperparameters}

\subsection{Decoding Hyperparameters}

The literal summarizers that generated the summaries which were then rescored using the answer and source reconstruction objected used the following decoding techniques:
\begin{itemize}
    \item Standard sampling (Temperature of 2)
    \item Nucleus sampling ($p = 0.95$, Temperature of 1.0)
    \item Nucleus sampling ($p = 0.95$, Temperature of 1.2)
    \item Nucleus sampling ($p = 0.95$, Temperature of 1.5)
    \item Nucleus sampling ($p = 0.95$, Temperature of 1.0)
\end{itemize}

\subsection{Fine-Tuning Hyperparameters}

We fine-tuned BART three times on each of the train-validation splits of the \multioped, \qmsum, and \squality datasets. The model was optimized using Adam with a weight decay of \(0.1\) and a linear scheduler with a warmup proportion of \(10\%\) of the total training time. We used a base batch size of 16 and performed a grid search over the learning rate, testing both \(3 \times 10^{-4}\) and \(3 \times 10^{-5}\), as well as the number of gradient accumulation steps, and thus the effective batch size, experimenting with values of \(1\), \(2\), and \(4\). Each model was trained for 50 epochs with early stopping set to 30 epochs.

\subsection{Rationality Parameters}

The values of the rationality parameters $\lambda$ and $\alpha$ for $\pragans,\pragsource,\pragas$ are computed \emph{dynamically} based on the order of magnitude of the likelihood scores. This is necessary because, for example, the negative log-likelihood score produced by the literal summarizer may be 40 while the one produced by the QA system might be 2500. As a result, the value of $\lambda$ is computed as
\[
    \lambda = c \times \lambda'
\]
where $c$ is a scaling constant computed as
\[
 c = 10^{-\frac{\texttt{max-order}}{\texttt{min-order}}}
\]
where $\texttt{max-order}$ is the order of the larger interpolated value (e.g., $2500 = 2.5 \times 10^3 \implies \texttt{max-order} = 3$) and $\texttt{min-order}$ is the order of the smaller interpolated value (e.g., $40 = 4 \times 10^1 \implies \texttt{min-order} = 1$). The range of values for $\lambda'$ include 
\begin{multline*}
\Lambda' \coloneq \{0.1, 0.2, \dots, 0.9, 1.0, 2.0, \dots, 9.0, \\ 10.0, 25.0, 50.0, 75.0, 100.0\}
\end{multline*}

\newpage

\onecolumn

\section{LLM Prompts}

\label{sec:llm_prompts}

In this section, we provide the prompts used for summary generation, question-answer generation as well as answer and source reconstruction objective prompts.

The prompt for summary generation was

\begin{quote}
    \textbf{System message:} You are a helpful, respectful and honest assistant specializing in query-focused summarization.
    
    \textbf{User message:} Read the following document and related questions. Provide a summary which both summarizes the document as well as answers the questions. That is, the summary should summarize the main points of the document while providing answers to the given questions. Do not generate anything else other than the summary. Questions: {q} Document: {s} Summary:
\end{quote}

The prompt for question-answer generation was

\begin{quote}
    \textbf{System message:} You are a helpful, respectful and honest assistant specializing in question answering.
    
    \textbf{User message:} Read the following document and then answer the question about it. Your answer should be clear, concise and accurate. Do not provide anything else other than the answer to the question. Document: {d} Question: {q} Answer:
\end{quote}

This prompt was also used to measure the answer reconstruction scores by replacing  with a generated summary and computing the likelihood of the corresponding answer.

The source reconstruction prompt was
\begin{quote}
    \textbf{System message:} You are a helpful, respectful and honest assistant specializing in the reconstruction of source documents given their summary.
    
    \textbf{User message:} Read the following summary and then regenerate the source document that it was about. Your regeneration should be as close to the original source document as possible. Do not provide anything else other than the reconstructed source document. Summary: {s} Source: 
\end{quote}

\section{Additional Results}

\todo{Can add the standard beam search and other SEAHORSE metrics results. But likely not necessary.}

\label{sec:app_add_results}

\subsection{Summarization Quality}

\subsubsection{Comparison of Alternative Pragmatic Summarizers with Summarization Metrics}

\label{sec:app_add_results_alternative_prags}

\todo{Missing standard beam search decoding.}

\begin{table}[H]
\centering
\resizebox{\linewidth}{!}{
\begin{tabular}{lcccccccccc}
\toprule
& \multicolumn{5}{c}{BART} & \multicolumn{5}{c}{Llama 3} \\
 & R-1 & R-2 & R-L & METEOR & BERTScore & R-1 & R-2 & R-L & METEOR & BERTScore \\
\cmidrule(r){2-6}\cmidrule(r){7-11}

 $\pragrandom$                    & 0.2963            & 0.0491            & 0.1545            & 0.2194               & 0.1365                  & 0.2982               & \textbf{0.0571}      & 0.1549               & 0.2624                  & 0.1336                     \\\cmidrule(r){2-6}\cmidrule(r){7-11}
 $\literalsumm$            & \textbf{0.3005}   & 0.0534            & \textbf{0.1603}   & 0.2140               & 0.1563                  & 0.2786               & 0.0567               & 0.1468               & 0.2749                  & 0.1236                     \\
 $\pragansonly$             & 0.2848            & 0.0447            & 0.1481            & 0.2110               & 0.1147                  & 0.2888               & 0.0538               & 0.1511               & 0.2549                  & 0.1201                     \\
 $\pragsourceonly$          & 0.2803            & 0.0434            & 0.1464            & 0.2116               & 0.1020                  & 0.2878               & 0.0530               & 0.1489               & 0.2608                  & 0.1230                     \\\cmidrule(r){2-6}\cmidrule(r){7-11}
 \optimalpragans             & 0.2962            & 0.0509            & 0.1562            & \textbf{0.2239}      & 0.1427                  & 0.3101               & 0.0562               & 0.1609               & 0.2616                  & 0.1530                     \\
 \optimalpragsource         & 0.3002            & \textbf{0.0535}   & 0.1602            & 0.2138               & \textbf{0.1563}         & 0.2953               & 0.0569               & 0.1530               & \textbf{0.2749}         & 0.1400                     \\
 \optimalpragas & 0.2979            & 0.0525            & 0.1595            & 0.2235               & 0.1541                  & \textbf{0.3105}      & 0.0567               & \textbf{0.1612}      & 0.2729                  & \textbf{0.1532}            \\\cmidrule(r){2-6}\cmidrule(r){7-11}
 $\pragoracle$                    & 0.3532            & 0.0812            & 0.1895            & 0.2732               & 0.2046                  & 0.3441               & 0.0838               & 0.1837               & 0.3056                  & 0.1972                     \\
\bottomrule
\end{tabular}
}
\caption{Performance of the different pragmatic summarizers on the \multioped dataset measured using the ROUGE, METEOR, and, BERTScore metrics. The summaries generated by the literal summarizer used standard sampling (temperature of 2).}
\label{tab:multioped_temp_2}
\end{table}

\begin{table}[H]
\centering
\resizebox{\linewidth}{!}{
\begin{tabular}{lcccccccccc}
\toprule
& \multicolumn{5}{c}{BART} & \multicolumn{5}{c}{Llama 3} \\
 & R-1 & R-2 & R-L & METEOR & BERTScore & R-1 & R-2 & R-L & METEOR & BERTScore \\
\cmidrule(r){2-6}\cmidrule(r){7-11}

 $\pragrandom$                    & 0.3435            & 0.0720            & 0.1798            & 0.2635               & 0.1994                  & 0.3115               & 0.0687               & 0.1684               & 0.2734                  & 0.1725                     \\\cmidrule(r){2-6}\cmidrule(r){7-11}
 $\literalsumm$            & 0.3440            & \textbf{0.0742}   & \textbf{0.1848}   & 0.2519               & 0.2091                  & 0.2913               & 0.0676               & 0.1610               & 0.2818                  & 0.1496                     \\
 $\pragansonly$             & 0.3383            & 0.0674            & 0.1778            & 0.2574               & 0.1887                  & 0.3081               & 0.0688               & 0.1678               & 0.2738                  & 0.1652                     \\
 $\pragsourceonly$          & 0.3376            & 0.0702            & 0.1781            & 0.2581               & 0.1881                  & 0.3092               & 0.0700               & 0.1688               & 0.2756                  & 0.1698                     \\\cmidrule(r){2-6}\cmidrule(r){7-11}
 \optimalpragans             & 0.3429            & 0.0724            & 0.1834            & \textbf{0.2639}      & 0.2046                  & 0.3314               & 0.0713               & \textbf{0.1791}      & 0.2799                  & 0.1993                     \\
 \optimalpragsource         & \textbf{0.3442}   & 0.0741            & 0.1848            & 0.2565               & \textbf{0.2092}         & 0.3119               & 0.0696               & 0.1698               & 0.2819                  & 0.1741                     \\
 \optimalpragas & 0.3429            & 0.0728            & 0.1837            & 0.2624               & 0.2073                  & \textbf{0.3317}      & \textbf{0.0717}      & 0.1791               & \textbf{0.2821}         & \textbf{0.1998}            \\\cmidrule(r){2-6}\cmidrule(r){7-11}
 $\pragoracle$                    & 0.3934            & 0.1052            & 0.2135            & 0.3093               & 0.2556                  & 0.3538               & 0.0945               & 0.1966               & 0.3128                  & 0.2274                     \\
\bottomrule
\end{tabular}
}
\caption{Performance of the different pragmatic summarizers on the \multioped dataset measured using the ROUGE, METEOR, and, BERTScore metrics. The summaries generated by the literal summarizer used nucleus sampling ($p = 0.95$, temperature of 1.0).}
\label{tab:multioped_nucleus_0.95}
\end{table}

\begin{table}[H]
\centering
\resizebox{\linewidth}{!}{
\begin{tabular}{lcccccccccc}
\toprule
& \multicolumn{5}{c}{BART} & \multicolumn{5}{c}{Llama 3} \\
 & R-1 & R-2 & R-L & METEOR & BERTScore & R-1 & R-2 & R-L & METEOR & BERTScore \\
\cmidrule(r){2-6}\cmidrule(r){7-11}

 $\pragrandom$                    & 0.3385            & 0.0674            & 0.1762            & 0.2579               & 0.1943                  & 0.3126               & 0.0707               & 0.1701               & 0.2741                  & 0.1730                     \\\cmidrule(r){2-6}\cmidrule(r){7-11}
 $\literalsumm$            & \textbf{0.3431}   & 0.0707            & \textbf{0.1812}   & 0.2489               & \textbf{0.2035}         & 0.2922               & 0.0670               & 0.1609               & 0.2817                  & 0.1520                     \\
 $\pragansonly$             & 0.3354            & 0.0669            & 0.1755            & 0.2536               & 0.1801                  & 0.3087               & 0.0704               & 0.1688               & 0.2734                  & 0.1657                     \\
 $\pragsourceonly$          & 0.3334            & 0.0660            & 0.1746            & 0.2513               & 0.1771                  & 0.3099               & 0.0705               & 0.1695               & 0.2756                  & 0.1725                     \\\cmidrule(r){2-6}\cmidrule(r){7-11}
 \optimalpragans             & 0.3425            & 0.0707            & 0.1799            & \textbf{0.2603}      & 0.1985                  & \textbf{0.3307}      & 0.0715               & 0.1791               & 0.2787                  & \textbf{0.1981}            \\
 \optimalpragsource         & 0.3430            & \textbf{0.0707}   & 0.1811            & 0.2526               & 0.2027                  & 0.3127               & 0.0708               & 0.1702               & 0.2819                  & 0.1785                     \\
 \optimalpragas & 0.3418            & 0.0707            & 0.1804            & 0.2595               & 0.2018                  & 0.3306               & \textbf{0.0715}      & \textbf{0.1791}      & \textbf{0.2819}         & 0.1981                     \\\cmidrule(r){2-6}\cmidrule(r){7-11}
 $\pragoracle$                    & 0.3922            & 0.1043            & 0.2122            & 0.3083               & 0.2495                  & 0.3552               & 0.0949               & 0.1975               & 0.3124                  & 0.2274                     \\
\bottomrule
\end{tabular}
}
\caption{Performance of the different pragmatic summarizers on the \multioped dataset measured using the ROUGE, METEOR, and, BERTScore metrics. The summaries generated by the literal summarizer used top-k sampling ($k = 640$, temperature of 1.0).}
\label{tab:multioped_topk_640}
\end{table}

\begin{table}[h]
\centering
\resizebox{\linewidth}{!}{
\begin{tabular}{lcccccccccc}
\toprule
& \multicolumn{5}{c}{BART} & \multicolumn{5}{c}{Llama 3} \\
 & R-1 & R-2 & R-L & METEOR & BERTScore & R-1 & R-2 & R-L & METEOR & BERTScore \\
\cmidrule(r){2-6}\cmidrule(r){7-11}

 $\pragrandom$                    & 0.3307            & 0.0638            & 0.1732            & 0.2498               & 0.1789                  & 0.3097               & 0.0664               & 0.1640               & 0.2714                  & 0.1712                     \\\cmidrule(r){2-6}\cmidrule(r){7-11}
 $\literalsumm$            & 0.3340            & 0.0682            & 0.1767            & 0.2413               & 0.1938                  & 0.2921               & 0.0655               & 0.1554               & 0.2840                  & 0.1551                     \\
 $\pragansonly$             & 0.3259            & 0.0622            & 0.1708            & 0.2430               & 0.1713                  & 0.3058               & 0.0654               & 0.1617               & 0.2702                  & 0.1613                     \\
 $\pragsourceonly$          & 0.3256            & 0.0607            & 0.1700            & 0.2443               & 0.1698                  & 0.3028               & 0.0660               & 0.1607               & 0.2718                  & 0.1673                     \\\cmidrule(r){2-6}\cmidrule(r){7-11}
 \optimalpragans            & 0.3294            & 0.0655            & 0.1746            & \textbf{0.2506}      & 0.1848                  & 0.3255               & \textbf{0.0682}      & 0.1716               & 0.2790                  & \textbf{0.1910}            \\
 \optimalpragsource          & \textbf{0.3342}   & \textbf{0.0689}   & \textbf{0.1772}   & 0.2450               & \textbf{0.1941}         & 0.3057               & 0.0661               & 0.1628               & \textbf{0.2842}         & 0.1744                     \\
 \optimalpragas & 0.3335            & 0.0687            & 0.1769            & 0.2500               & 0.1927                  & \textbf{0.3258}      & 0.0679               & \textbf{0.1718}      & 0.2839                  & 0.1910                     \\\cmidrule(r){2-6}\cmidrule(r){7-11}
 $\pragoracle$                    & 0.3835            & 0.0978            & 0.2072            & 0.2989               & 0.2421                  & 0.3560               & 0.0938               & 0.1937               & 0.3145                  & 0.2274                     \\
\bottomrule
\end{tabular}
}
\caption{Performance of the different pragmatic summarizers on the \multioped dataset measured using the ROUGE, METEOR, and BERTScore metrics. The summaries generated by the literal summarizer used nucleus sampling ($p = 0.95$, temperature of 1.5).}
\label{tab:multioped_nucleus_0.95_temp_1.5}
\end{table}

\begin{table}[H]
\centering
\resizebox{\linewidth}{!}{
\begin{tabular}{lcccccccccc}
\toprule
& \multicolumn{5}{c}{BART} & \multicolumn{5}{c}{Llama 3} \\
 & R-1 & R-2 & R-L & METEOR & BERTScore & R-1 & R-2 & R-L & METEOR & BERTScore \\
\cmidrule(r){2-6}\cmidrule(r){7-11}

 $\pragrandom$                    & \textbf{0.2499}   & 0.0345            & 0.1392            & 0.2198               & 0.1023                  & 0.2754               & 0.0720               & 0.1673               & 0.3067                  & 0.1286                     \\\cmidrule(r){2-6}\cmidrule(r){7-11}
 $\literalsumm$            & 0.2224            & 0.0349            & 0.1396            & 0.1580               & 0.1208                  & 0.2421               & 0.0678               & 0.1480               & 0.3069                  & 0.0806                     \\
 $\pragansonly$             & 0.2244            & 0.0286            & 0.1305            & 0.1896               & 0.0930                  & 0.2719               & 0.0685               & 0.1636               & 0.2998                  & 0.1231                     \\
 $\pragsourceonly$          & 0.2432            & 0.0329            & 0.1338            & 0.2304               & 0.0896                  & 0.2675               & 0.0697               & 0.1618               & 0.3041                  & 0.1179                     \\\cmidrule(r){2-6}\cmidrule(r){7-11}
 \optimalpragans             & 0.2433            & 0.0349            & 0.1393            & \textbf{0.2622}      & 0.1199                  & 0.3040               & 0.0749               & 0.1841               & 0.3055                  & 0.1795                     \\
 \optimalpragsource         & 0.2433            & 0.0349            & 0.1396            & 0.2497               & 0.1208                  & 0.2984               & 0.0751               & 0.1809               & 0.3082                  & 0.1708                     \\
 \optimalpragas & 0.2487            & \textbf{0.0355}   & \textbf{0.1414}   & 0.2609               & \textbf{0.1209}         & \textbf{0.3040}      & \textbf{0.0755}      & \textbf{0.1843}      & \textbf{0.3086}         & \textbf{0.1796}            \\\cmidrule(r){2-6}\cmidrule(r){7-11}
 $\pragoracle$                    & 0.3299            & 0.0822            & 0.1946            & 0.3131               & 0.1886                  & 0.3388               & 0.1127               & 0.2134               & 0.3721                  & 0.2221                     \\
\bottomrule
\end{tabular}
}
\caption{Performance of the different pragmatic summarizers on the \qmsum dataset measured using the ROUGE, METEOR, and, BERTScore metrics. The summaries generated by the literal summarizer used standard sampling (temperature of 2).}
\label{tab:qmsum_temp_2}
\end{table}

\begin{table}[H]
\centering
\resizebox{\linewidth}{!}{
\begin{tabular}{lcccccccccc}
\toprule
& \multicolumn{5}{c}{BART} & \multicolumn{5}{c}{Llama 3} \\
 & R-1 & R-2 & R-L & METEOR & BERTScore & R-1 & R-2 & R-L & METEOR & BERTScore \\
\cmidrule(r){2-6}\cmidrule(r){7-11}

 $\pragrandom$                    & 0.3846            & 0.1412            & 0.2507            & \textbf{0.3228}      & 0.2413                  & 0.3023               & 0.0984               & 0.1954               & 0.3363                  & 0.1772                     \\\cmidrule(r){2-6}\cmidrule(r){7-11}
 $\literalsumm$            & 0.3815            & 0.1558            & 0.2690            & 0.2899               & 0.2702                  & 0.2710               & 0.0916               & 0.1788               & 0.3339                  & 0.1233                     \\
 $\pragansonly$             & 0.3627            & 0.1217            & 0.2326            & 0.2856               & 0.2213                  & 0.2973               & 0.0936               & 0.1925               & 0.3291                  & 0.1613                     \\
 $\pragsourceonly$          & 0.3731            & 0.1295            & 0.2384            & 0.3121               & 0.2246                  & 0.2952               & 0.0967               & 0.1911               & 0.3339                  & 0.1577                     \\\cmidrule(r){2-6}\cmidrule(r){7-11}
 \optimalpragans             & 0.3812            & 0.1557            & 0.2686            & 0.3141               & 0.2697                  & 0.3253               & 0.1002               & 0.2095               & 0.3344                  & \textbf{0.2111}            \\
 \optimalpragsource         & 0.3828            & 0.1566            & 0.2696            & 0.3125               & 0.2711                  & 0.3130               & 0.0988               & 0.2016               & 0.3343                  & 0.1902                     \\
 \optimalpragas & \textbf{0.3860}   & \textbf{0.1574}   & \textbf{0.2704}   & 0.3203               & \textbf{0.2722}         & \textbf{0.3255}      & \textbf{0.1004}      & \textbf{0.2097}      & \textbf{0.3366}         & 0.2109                     \\\cmidrule(r){2-6}\cmidrule(r){7-11}
 $\pragoracle$                    & 0.4836            & 0.2346            & 0.3442            & 0.4320               & 0.3490                  & 0.3545               & 0.1372               & 0.2384               & 0.3979                  & 0.2502                     \\
\bottomrule
\end{tabular}
}
\caption{Performance of the different pragmatic summarizers on the \qmsum dataset measured using the ROUGE, METEOR, and, BERTScore metrics. The summaries generated by the literal summarizer used nucleus sampling ($p = 0.95$, temperature of 1.0).}
\label{tab:qmsum_nucleus_0.95}
\end{table}

\begin{table}[H]
\centering
\resizebox{\linewidth}{!}{
\begin{tabular}{lcccccccccc}
\toprule
& \multicolumn{5}{c}{BART} & \multicolumn{5}{c}{Llama 3} \\
 & R-1 & R-2 & R-L & METEOR & BERTScore & R-1 & R-2 & R-L & METEOR & BERTScore \\
\cmidrule(r){2-6}\cmidrule(r){7-11}

 $\pragrandom$                    & 0.3658            & 0.1254            & 0.2352            & 0.3051               & 0.2245                  & 0.3013               & 0.0986               & 0.1952               & 0.3352                  & 0.1766                     \\\cmidrule(r){2-6}\cmidrule(r){7-11}
 $\literalsumm$            & 0.3669            & 0.1424            & 0.2547            & 0.2753               & 0.2518                  & 0.2727               & 0.0930               & 0.1798               & 0.3343                  & 0.1244                     \\
 $\pragansonly$             & 0.3372            & 0.1046            & 0.2181            & 0.2656               & 0.1964                  & 0.2979               & 0.0955               & 0.1931               & 0.3307                  & 0.1620                     \\
 $\pragsourceonly$          & 0.3558            & 0.1147            & 0.2251            & 0.2977               & 0.2074                  & 0.2964               & 0.0964               & 0.1924               & 0.3337                  & 0.1610                     \\\cmidrule(r){2-6}\cmidrule(r){7-11}
 \optimalpragans             & 0.3644            & 0.1405            & 0.2531            & 0.2985               & 0.2484                  & 0.3279               & \textbf{0.1030}      & \textbf{0.2107}      & 0.3344                  & 0.2130                     \\
 \optimalpragsource         & 0.3670            & 0.1430            & 0.2555            & 0.3040               & 0.2519                  & 0.3161               & 0.1015               & 0.2043               & 0.3344                  & 0.1957                     \\
 \optimalpragas & \textbf{0.3707}   & \textbf{0.1434}   & \textbf{0.2560}   & \textbf{0.3094}      & \textbf{0.2523}         & \textbf{0.3282}      & 0.1030               & 0.2106               & \textbf{0.3355}         & \textbf{0.2133}            \\\cmidrule(r){2-6}\cmidrule(r){7-11}
 $\pragoracle$                    & 0.4673            & 0.2183            & 0.3281            & 0.4150               & 0.3305                  & 0.3547               & 0.1367               & 0.2380               & 0.3954                  & 0.2494                     \\
\bottomrule
\end{tabular}
}
\caption{Performance of the different pragmatic summarizers on the \qmsum dataset measured using the ROUGE, METEOR, and, BERTScore metrics. The summaries generated by the literal summarizer used top-k sampling ($k = 640$, temperature of 1.0).}
\label{tab:qmsum_topk_640}
\end{table}

\begin{table}[H]
\centering
\resizebox{\linewidth}{!}{
\begin{tabular}{lcccccccccc}
\toprule
& \multicolumn{5}{c}{BART} & \multicolumn{5}{c}{Llama 3} \\
 & R-1 & R-2 & R-L & METEOR & BERTScore & R-1 & R-2 & R-L & METEOR & BERTScore \\
\cmidrule(r){2-6}\cmidrule(r){7-11}

 $\pragrandom$                    & 0.3652            & 0.1221            & 0.2308            & 0.3060               & 0.2227                  & 0.2928               & 0.0918               & 0.1874               & \textbf{0.3307}         & 0.1622                     \\\cmidrule(r){2-6}\cmidrule(r){7-11}
 $\literalsumm$            & 0.3585            & 0.1334            & 0.2456            & 0.2677               & 0.2477                  & 0.2620               & 0.0864               & 0.1707               & 0.3284                  & 0.1069                     \\
 $\pragansonly$             & 0.3380            & 0.1019            & 0.2120            & 0.2684               & 0.1984                  & 0.2890               & 0.0873               & 0.1838               & 0.3217                  & 0.1462                     \\
 $\pragsourceonly$          & 0.3524            & 0.1078            & 0.2187            & 0.2962               & 0.2032                  & 0.2891               & 0.0914               & 0.1853               & 0.3281                  & 0.1475                     \\\cmidrule(r){2-6}\cmidrule(r){7-11}
 \optimalpragans             & 0.3571            & 0.1314            & 0.2445            & 0.3063               & 0.2456                  & 0.3240               & 0.0963               & 0.2045               & 0.3280                  & 0.2063                     \\
 \optimalpragsource         & 0.3599            & 0.1332            & 0.2455            & 0.3015               & 0.2474                  & 0.3116               & 0.0956               & 0.1982               & 0.3284                  & 0.1866                     \\
 \optimalpragas & \textbf{0.3674}   & \textbf{0.1346}   & \textbf{0.2473}   & \textbf{0.3083}      & \textbf{0.2498}         & \textbf{0.3242}      & \textbf{0.0963}      & \textbf{0.2046}      & 0.3298                  & \textbf{0.2065}            \\\cmidrule(r){2-6}\cmidrule(r){7-11}
 $\pragoracle$                    & 0.4594            & 0.2072            & 0.3183            & 0.4134               & 0.3249                  & 0.3546               & 0.1329               & 0.2345               & 0.3947                  & 0.2459                     \\
\bottomrule
\end{tabular}
}
\caption{Performance of the different pragmatic summarizers on the \qmsum dataset measured using the ROUGE, METEOR, and, BERTScore metrics. The summaries generated by the literal summarizer used nucleus sampling ($p = 0.95$, temperature of 1.2).}
\label{tab:qmsum_nucleus_0.95_temp_1.2}
\end{table}

\begin{table}[H]
\centering
\resizebox{\linewidth}{!}{
\begin{tabular}{lcccccccccc}
\toprule
& \multicolumn{5}{c}{BART} & \multicolumn{5}{c}{Llama 3} \\
 & R-1 & R-2 & R-L & METEOR & BERTScore & R-1 & R-2 & R-L & METEOR & BERTScore \\
\cmidrule(r){2-6}\cmidrule(r){7-11}

 $\pragrandom$                    & \textbf{0.3225}   & 0.0837            & 0.1939            & 0.2690               & 0.1842                  & 0.2900               & 0.0859               & 0.1815               & 0.3224                  & 0.1577                     \\\cmidrule(r){2-6}\cmidrule(r){7-11}
 $\literalsumm$            & 0.3116            & \textbf{0.0962}   & 0.2071            & 0.2295               & 0.2076                  & 0.2559               & 0.0806               & 0.1631               & 0.3207                  & 0.0982                     \\
 $\pragansonly$             & 0.2955            & 0.0672            & 0.1783            & 0.2367               & 0.1598                  & 0.2832               & 0.0805               & 0.1773               & 0.3156                  & 0.1362                     \\
 $\pragsourceonly$          & 0.3119            & 0.0740            & 0.1820            & 0.2693               & 0.1690                  & 0.2840               & 0.0837               & 0.1778               & 0.3204                  & 0.1412                     \\\cmidrule(r){2-6}\cmidrule(r){7-11}
 \optimalpragans             & 0.3098            & 0.0943            & 0.2053            & 0.2890               & 0.2048                  & \textbf{0.3192}      & \textbf{0.0895}      & 0.1991               & 0.3203                  & \textbf{0.2011}            \\
 \optimalpragsource         & 0.3116            & 0.0960            & 0.2069            & 0.2815               & 0.2071                  & 0.3077               & 0.0883               & 0.1933               & 0.3221                  & 0.1843                     \\
 \optimalpragas & 0.3194            & 0.0960            & \textbf{0.2081}   & \textbf{0.2892}      & \textbf{0.2082}         & 0.3189               & 0.0893               & \textbf{0.1992}      & \textbf{0.3234}         & 0.2010                     \\\cmidrule(r){2-6}\cmidrule(r){7-11}
 $\pragoracle$                    & 0.4173            & 0.1593            & 0.2719            & 0.3783               & 0.2802                  & 0.3493               & 0.1270               & 0.2276               & 0.3893                  & 0.2390                     \\
\bottomrule
\end{tabular}
}
\caption{Performance of the different pragmatic summarizers on the \qmsum dataset measured using the ROUGE, METEOR, and, BERTScore metrics. The summaries generated by the literal summarizer used nucleus sampling ($p = 0.95$, temperature of 1.5).}
\label{tab:qmsum_nucleus_0.95_temp_1.5}
\end{table}

\begin{table}[H]
\centering
\resizebox{\linewidth}{!}{
\begin{tabular}{lcccccccccc}
\toprule
& \multicolumn{5}{c}{BART} & \multicolumn{5}{c}{Llama 3} \\
 & R-1 & R-2 & R-L & METEOR & BERTScore & R-1 & R-2 & R-L & METEOR & BERTScore \\
\cmidrule(r){2-6}\cmidrule(r){7-11}

 $\pragrandom$                    & 0.2915            & 0.0310            & 0.1329            & 0.2153               & -0.0056                 & 0.3541               & 0.0705               & 0.1730               & 0.2554                  & 0.0951                     \\\cmidrule(r){2-6}\cmidrule(r){7-11}
 $\literalsumm$            & 0.2327            & 0.0245            & 0.1225            & 0.1426               & -0.0108                 & 0.3753               & \textbf{0.0798}      & 0.1796               & 0.2718                  & 0.0975                     \\
 $\pragansonly$             & 0.2681            & 0.0279            & 0.1275            & 0.1895               & -0.0154                 & 0.3468               & 0.0679               & 0.1738               & 0.2393                  & 0.0865                     \\
 $\pragsourceonly$          & 0.3018            & \textbf{0.0319}   & \textbf{0.1342}   & 0.2267               & -0.0055                 & 0.3569               & 0.0643               & 0.1685               & 0.2567                  & 0.0731                     \\\cmidrule(r){2-6}\cmidrule(r){7-11}
 \optimalpragans             & \textbf{0.3039}   & 0.0294            & 0.1316            & \textbf{0.2318}      & \textbf{0.0000}         & 0.3694               & 0.0794               & \textbf{0.1800}      & 0.2631                  & \textbf{0.0989}            \\
 \optimalpragsource         & 0.2863            & 0.0305            & 0.1305            & 0.2094               & \textbf{0.0000}         & \textbf{0.3756}      & \textbf{0.0798}      & 0.1796               & \textbf{0.2719}         & 0.0975                     \\
 \optimalpragas & 0.3028            & 0.0306            & 0.1323            & 0.2305               & \textbf{0.0000}         & 0.3748               & 0.0793               & 0.1792               & 0.2716                  & 0.0974                     \\\cmidrule(r){2-6}\cmidrule(r){7-11}
 $\pragoracle$                    & 0.3442            & 0.0480            & 0.1530            & 0.2594               & 0.0377                  & 0.4082               & 0.1022               & 0.2039               & 0.2937                  & 0.1570                     \\
\bottomrule
\end{tabular}
}
\caption{Performance of the different pragmatic summarizers on the \squality dataset measured using the ROUGE, METEOR, and, BERTScore metrics. The summaries generated by the literal summarizer used standard sampling (temperature of 2).}
\label{tab:squality_temp_2}
\end{table}

\begin{table}[H]
\centering
\resizebox{\linewidth}{!}{
\begin{tabular}{lcccccccccc}
\toprule
& \multicolumn{5}{c}{BART} & \multicolumn{5}{c}{Llama 3} \\
 & R-1 & R-2 & R-L & METEOR & BERTScore & R-1 & R-2 & R-L & METEOR & BERTScore \\
\cmidrule(r){2-6}\cmidrule(r){7-11}

 $\pragrandom$                    & 0.3763            & \textbf{0.0780}   & \textbf{0.1837}   & 0.2466               & 0.0713                  & 0.3944               & 0.1000               & 0.2018               & 0.2800                  & 0.1257                     \\\cmidrule(r){2-6}\cmidrule(r){7-11}
 $\literalsumm$            & 0.3307            & 0.0696            & 0.1774            & 0.1955               & 0.0717                  & 0.3982               & 0.1013               & 0.2005               & 0.2859                  & 0.1117                     \\
 $\pragansonly$             & 0.3539            & 0.0693            & 0.1789            & 0.2294               & 0.0622                  & 0.3923               & 0.0984               & 0.2016               & 0.2746                  & 0.1135                     \\
 $\pragsourceonly$          & 0.3749            & 0.0744            & 0.1816            & 0.2576               & 0.0647                  & \textbf{0.4014}      & 0.1022               & 0.2023               & 0.2859                  & 0.1227                     \\\cmidrule(r){2-6}\cmidrule(r){7-11}
 \optimalpragans             & \textbf{0.3778}   & 0.0739            & 0.1798            & \textbf{0.2651}      & 0.0719                  & 0.3985               & 0.1032               & \textbf{0.2066}      & 0.2861                  & \textbf{0.1423}            \\
 \optimalpragsource         & 0.3551            & 0.0718            & 0.1799            & 0.2309               & \textbf{0.0724}         & 0.4011               & 0.1021               & 0.2015               & \textbf{0.2886}         & 0.1209                     \\
 \optimalpragas & 0.3768            & 0.0750            & 0.1818            & 0.2626               & 0.0720                  & 0.3999               & \textbf{0.1041}      & 0.2058               & 0.2875                  & 0.1329                     \\\cmidrule(r){2-6}\cmidrule(r){7-11}
 $\pragoracle$                    & 0.4212            & 0.1031            & 0.2076            & 0.2909               & 0.1183                  & 0.4381               & 0.1297               & 0.2285               & 0.3145                  & 0.1777                     \\
\bottomrule
\end{tabular}
}
\caption{Performance of the different pragmatic summarizers on the \squality dataset measured using the ROUGE, METEOR, and, BERTScore metrics. The summaries generated by the literal summarizer used nucleus sampling ($p = 0.95$, temperature of 1.0).}
\label{tab:squality_nucleus_0.95}
\end{table}

\begin{table}[H]
\centering
\resizebox{\linewidth}{!}{
\begin{tabular}{lcccccccccc}
\toprule
& \multicolumn{5}{c}{BART} & \multicolumn{5}{c}{Llama 3} \\
 & R-1 & R-2 & R-L & METEOR & BERTScore & R-1 & R-2 & R-L & METEOR & BERTScore \\
\cmidrule(r){2-6}\cmidrule(r){7-11}

 $\pragrandom$                    & 0.3452            & 0.0626            & 0.1702            & 0.2317               & \textbf{0.0547}         & 0.3951               & 0.1012               & 0.2016               & 0.2807                  & 0.1246                     \\\cmidrule(r){2-6}\cmidrule(r){7-11}
 $\literalsumm$            & 0.2979            & 0.0548            & 0.1628            & 0.1774               & 0.0517                  & 0.3997               & 0.1009               & 0.1990               & 0.2880                  & 0.1124                     \\
 $\pragansonly$             & 0.3254            & 0.0561            & 0.1649            & 0.2123               & 0.0448                  & 0.3928               & 0.0964               & 0.2001               & 0.2736                  & 0.1104                     \\
 $\pragsourceonly$          & 0.3557            & \textbf{0.0635}   & \textbf{0.1717}   & 0.2496               & 0.0536                  & \textbf{0.4012}      & 0.1022               & 0.2018               & 0.2838                  & 0.1223                     \\\cmidrule(r){2-6}\cmidrule(r){7-11}
 \optimalpragans             & \textbf{0.3558}   & 0.0621            & 0.1688            & \textbf{0.2550}      & 0.0515                  & 0.3989               & 0.1011               & 0.2052               & 0.2873                  & \textbf{0.1403}            \\
 \optimalpragsource         & 0.3296            & 0.0589            & 0.1678            & 0.2199               & 0.0536                  & 0.4012               & \textbf{0.1029}      & 0.2013               & \textbf{0.2901}         & 0.1211                     \\
 \optimalpragas & 0.3552            & 0.0626            & 0.1703            & 0.2530               & 0.0522                  & 0.4011               & 0.1027               & \textbf{0.2053}      & 0.2884                  & 0.1330                     \\\cmidrule(r){2-6}\cmidrule(r){7-11}
 $\pragoracle$                    & 0.3984            & 0.0889            & 0.1961            & 0.2804               & 0.1032                  & 0.4392               & 0.1294               & 0.2264               & 0.3142                  & 0.1778                     \\
\bottomrule
\end{tabular}
}
\caption{Performance of the different pragmatic summarizers on the \squality dataset measured using the ROUGE, METEOR, and, BERTScore metrics. The summaries generated by the literal summarizer used top-k sampling ($k = 640$, temperature of 1.0).}
\label{tab:squality_topk_640}
\end{table}

\begin{table}[H]
\centering
\resizebox{\linewidth}{!}{
\begin{tabular}{lcccccccccc}
\toprule
& \multicolumn{5}{c}{BART} & \multicolumn{5}{c}{Llama 3} \\
 & R-1 & R-2 & R-L & METEOR & BERTScore & R-1 & R-2 & R-L & METEOR & BERTScore \\
\cmidrule(r){2-6}\cmidrule(r){7-11}

 $\pragrandom$                    & 0.3543            & 0.0633            & \textbf{0.1723}   & 0.2408               & \textbf{0.0600}         & 0.3907               & 0.0943               & 0.1959               & 0.2731                  & 0.1231                     \\\cmidrule(r){2-6}\cmidrule(r){7-11}
 $\literalsumm$            & 0.3029            & 0.0543            & 0.1623            & 0.1796               & 0.0553                  & 0.3980               & 0.0975               & 0.1977               & 0.2863                  & 0.1097                     \\
 $\pragansonly$             & 0.3374            & 0.0588            & 0.1678            & 0.2182               & 0.0501                  & 0.3830               & 0.0903               & 0.1943               & 0.2625                  & 0.1073                     \\
 $\pragsourceonly$          & 0.3650            & \textbf{0.0651}   & 0.1712            & 0.2558               & 0.0558                  & 0.3971               & 0.0988               & 0.1976               & 0.2800                  & 0.1230                     \\\cmidrule(r){2-6}\cmidrule(r){7-11}
 \optimalpragans             & \textbf{0.3670}   & 0.0631            & 0.1692            & \textbf{0.2614}      & 0.0556                  & 0.3975               & 0.0986               & \textbf{0.2000}      & 0.2858                  & \textbf{0.1333}            \\
 \optimalpragsource         & 0.3378            & 0.0603            & 0.1668            & 0.2238               & 0.0565                  & \textbf{0.3991}      & 0.1002               & 0.1980               & 0.2876                  & 0.1225                     \\
 \optimalpragas & 0.3660            & 0.0643            & 0.1709            & 0.2592               & 0.0559                  & 0.3987               & \textbf{0.1008}      & 0.1993               & \textbf{0.2878}         & 0.1302                     \\\cmidrule(r){2-6}\cmidrule(r){7-11}
 $\pragoracle$                    & 0.4069            & 0.0895            & 0.1962            & 0.2863               & 0.1043                  & 0.4372               & 0.1252               & 0.2240               & 0.3118                  & 0.1788                     \\
\bottomrule
\end{tabular}
}
\caption{Performance of the different pragmatic summarizers on the \squality dataset measured using the ROUGE, METEOR, and, BERTScore metrics. The summaries generated by the literal summarizer used nucleus sampling ($p = 0.95$, temperature of 1.2).}
\label{tab:squality_nucleus_0.95_temp_1.2}
\end{table}

\begin{table}[H]
\centering
\resizebox{\linewidth}{!}{
\begin{tabular}{lcccccccccc}
\toprule
& \multicolumn{5}{c}{BART} & \multicolumn{5}{c}{Llama 3} \\
 & R-1 & R-2 & R-L & METEOR & BERTScore & R-1 & R-2 & R-L & METEOR & BERTScore \\
\cmidrule(r){2-6}\cmidrule(r){7-11}

 $\pragrandom$                    & 0.3337            & 0.0489            & 0.1561            & 0.2277               & 0.0358                  & 0.3808               & 0.0882               & 0.1913               & 0.2679                  & 0.1198                     \\\cmidrule(r){2-6}\cmidrule(r){7-11}
 $\literalsumm$            & 0.2701            & 0.0412            & 0.1449            & 0.1588               & 0.0309                  & 0.3948               & 0.0942               & 0.1936               & 0.2840                  & 0.1100                     \\
 $\pragansonly$             & 0.3109            & 0.0451            & 0.1507            & 0.2107               & 0.0308                  & 0.3793               & 0.0873               & 0.1919               & 0.2606                  & 0.1040                     \\
 $\pragsourceonly$          & 0.3394            & \textbf{0.0491}   & \textbf{0.1561}   & 0.2456               & \textbf{0.0367}         & 0.3882               & 0.0893               & 0.1911               & 0.2710                  & 0.1168                     \\\cmidrule(r){2-6}\cmidrule(r){7-11}
 \optimalpragans             & \textbf{0.3433}   & 0.0483            & 0.1540            & \textbf{0.2526}      & 0.0312                  & 0.3948               & 0.0942               & \textbf{0.1951}      & 0.2836                  & \textbf{0.1279}            \\
 \optimalpragsource         & 0.3114            & 0.0459            & 0.1507            & 0.2126               & 0.0325                  & 0.3956               & \textbf{0.0944}      & 0.1936               & \textbf{0.2844}         & 0.1176                     \\
 \optimalpragas & 0.3400            & 0.0488            & 0.1544            & 0.2482               & 0.0345                  & \textbf{0.3959}      & 0.0943               & 0.1946               & 0.2844                  & 0.1256                     \\\cmidrule(r){2-6}\cmidrule(r){7-11}
 $\pragoracle$                    & 0.3850            & 0.0713            & 0.1800            & 0.2769               & 0.0816                  & 0.4305               & 0.1210               & 0.2190               & 0.3083                  & 0.1741                     \\
\bottomrule
\end{tabular}
}
\caption{Performance of the different pragmatic summarizers on the \squality dataset measured using the ROUGE, METEOR, and, BERTScore metrics. The summaries generated by the literal summarizer used nucleus sampling ($p = 0.95$, temperature of 1.5).}
\label{tab:squality_nucleus_0.95_temp_1.5}
\end{table}

\subsubsection{Pragmatic Summarizer Top Score Frequencies}

\label{sec:app_prag_summ_top_score_freq}

\begin{table}[H]
\centering
\resizebox{0.5\linewidth}{!}{
\begin{tabular}{lccccc}
\toprule
                                  &   R-1 &   R-2 &   R-L &   METEOR &   BERTScore \\
\cmidrule(r){2-6}
 $\pragrandom$                    &     2 &     0 &     0 &        2 &           0 \\\cmidrule(r){2-6}
 $\literalsumm$            &     0 &     1 &     0 &        0 &           0 \\\cmidrule(r){2-6}
 $\pragansonly$             &     0 &     0 &     0 &        0 &           0 \\
 $\pragsourceonly$          &     0 &     0 &     0 &        0 &           0 \\
 \optimalpragans             &     1 &     2 &     1 &        1 &           2 \\
 \optimalpragsource         &     0 &     0 &     0 &        0 &           0 \\
 \optimalpragas &     7 &     7 &     9 &        7 &           8 \\
\bottomrule
\end{tabular}
}
\caption{Number of times each pragmatic summarizer achieves the highest ROUGE, METEOR, and BERTScore scores on the \qmsum dataset. We aggregate counts across all combinations of models (Llama 3 and BART) and decoding types (Standard, nucleus, and top-k sampling).}
\label{tab:qmsum_prag_summ_top_score_freq}
\end{table}

\begin{table}[H]
\centering
\resizebox{0.5\linewidth}{!}{
\begin{tabular}{lccccc}
\toprule
                                  &   R-1 &   R-2 &   R-L &   METEOR &   BERTScore \\
\cmidrule(r){2-6}
 $\pragrandom$                    &     0 &     1 &     2 &        0 &           2 \\\cmidrule(r){2-6}
 $\literalsumm$            &     0 &     1 &     0 &        0 &           0 \\\cmidrule(r){2-6}
 $\pragansonly$             &     0 &     0 &     0 &        0 &           0 \\
 $\pragsourceonly$          &     2 &     4 &     3 &        0 &           1 \\
 \optimalpragans             &     5 &     0 &     4 &        5 &           6 \\
 \optimalpragsource         &     2 &     2 &     0 &        4 &           1 \\
 \optimalpragas &     1 &     2 &     1 &        1 &           0 \\
\bottomrule
\end{tabular}
}
\caption{Number of times each pragmatic summarizer achieves the highest ROUGE, METEOR, and BERTScore scores on the \squality dataset. We aggregate counts across all combinations of models (Llama 3 and BART) and decoding types (Standard, nucleus, and top-k sampling).}
\label{tab:squality_prag_summ_top_score_freq}
\end{table}

\subsubsection{Summarization Metric Score Changes Relative to $\pragrandom$}

\label{sec:app_relative_change_scores}

\begin{table}[H]
\centering
\resizebox{1\linewidth}{!}{
\begin{tabular}{lcccccccccc}
\toprule
& \multicolumn{5}{c}{BART} & \multicolumn{5}{c}{Llama 3} \\
 & R-1 & R-2 & R-L & METEOR & BERTScore & R-1 & R-2 & R-L & METEOR & BERTScore \\
\cmidrule(r){2-6}\cmidrule(r){7-11}

 $\literalsumm$            & $+$0.20             & $+$1.38             & $+$0.41             & $-$\underline{0.69}  & $+$1.66                 & $-$\underline{1.14}  & $-$0.29              & $-$\underline{1.04}  & $+$0.93                 & $-$\underline{1.88}        \\\cmidrule(r){2-6}\cmidrule(r){7-11}
 $\pragansonly$             & $-$0.29             & $-$0.51             & $-$0.27             & $-$0.54              & $-$0.85                 & $-$0.25              & $-$\underline{0.31}  & $-$0.27              & $-$\underline{0.09}     & $-$1.16                    \\
 $\pragsourceonly$          & $-$\underline{0.31} & $-$\underline{0.97} & $-$\underline{0.37} & $-$0.44              & $-$\underline{1.02}     & $-$0.44              & $-$0.13              & $-$0.39              & $+$0.03                 & $-$0.46                    \\\cmidrule(r){2-6}\cmidrule(r){7-11}
 \optimalpragans             & $-$0.08             & $+$0.53             & $+$0.16             & $+$\textbf{0.06}     & $+$0.66                 & $+$1.02              & $+$\textbf{0.52}     & $+$0.93              & $+$0.55                 & $+$\textbf{2.31}           \\
 \optimalpragsource         & $+$\textbf{0.21}    & $+$\textbf{1.58}    & $+$\textbf{0.47}    & $-$0.39              & $+$\textbf{1.70}        & $-$0.26              & $-$0.09              & $-$0.15              & $+$\textbf{0.94}        & $+$0.37                    \\
 \optimalpragas & $+$0.17             & $+$1.53             & $+$0.43             & $+$0.01              & $+$1.54                 & $+$\textbf{1.04}     & $+$0.43              & $+$\textbf{0.95}     & $+$0.92                 & $+$2.30                    \\
\bottomrule
\end{tabular}
}
\caption{Relative change in ROUGE, METEOR, and BERTScore scores between the pragmatic summarizers and $\pragrandom$ on the \multioped dataset. Relative changes are aggregated across all decoding methods. A value of $+x$ ($-x$) indicates that the average metric score achieved is $x\%$ higher (lower) than the one achieved by $\pragrandom$. The lowest relative change is \underline{underlined}, and the highest is \textbf{bolded}.}
\label{tab:multioped_relative_to_random_scores}
\end{table}

\begin{table}[H]
\centering
\resizebox{1\linewidth}{!}{
\begin{tabular}{lcccccccccc}
\toprule
& \multicolumn{5}{c}{BART} & \multicolumn{5}{c}{Llama 3} \\
 & R-1 & R-2 & R-L & METEOR & BERTScore & R-1 & R-2 & R-L & METEOR & BERTScore \\
\cmidrule(r){2-6}\cmidrule(r){7-11}

 $\literalsumm$            & $-$0.67             & $+$\textbf{2.98}    & $+$1.36             & $-$\underline{2.93}  & $+$2.53                 & $-$\underline{2.35}  & $-$1.23              & $-$\underline{2.03}  & $-$0.11                 & $-$\underline{7.55}        \\\cmidrule(r){2-6}\cmidrule(r){7-11}
 $\pragansonly$             & $-$\underline{1.67} & $-$\underline{3.94} & $-$\underline{1.61} & $-$2.40              & $-$\underline{2.65}     & $-$0.46              & $-$\underline{1.25}  & $-$0.46              & $-$\underline{0.42}     & $-$2.73                    \\
 $\pragsourceonly$          & $-$0.66             & $-$2.30             & $-$1.23             & $+$0.03              & $-$1.65                 & $-$0.41              & $-$0.50              & $-$0.41              & $-$0.12                 & $-$2.10                    \\\cmidrule(r){2-6}\cmidrule(r){7-11}
 \optimalpragans             & $-$0.79             & $+$2.54             & $+$1.18             & $+$1.49              & $+$2.23                 & $+$\textbf{2.02}     & $+$\textbf{0.86}     & $+$1.94              & $-$0.13                 & $+$\textbf{5.50}           \\
 \optimalpragsource         & $-$0.68             & $+$2.94             & $+$1.34             & $+$0.94              & $+$2.48                 & $+$1.23              & $+$0.57              & $+$1.30              & $-$0.02                 & $+$3.37                    \\
 \optimalpragas & $-$\textbf{0.19}    & $+$2.94             & $+$\textbf{1.47}    & $+$\textbf{1.50}     & $+$\textbf{2.60}        & $+$2.00              & $+$0.80              & $+$\textbf{1.95}     & $+$\textbf{0.07}        & $+$5.49                    \\
\bottomrule
\end{tabular}
}
\caption{Relative change in ROUGE, METEOR, and BERTScore scores between the pragmatic summarizers and $\pragrandom$ on the \qmsum dataset. Relative changes are aggregated across all decoding methods. A value of $+x$ ($-x$) indicates that the average metric score achieved is $x\%$ higher (lower) than the one achieved by $\pragrandom$. The lowest relative change is \underline{underlined}, and the highest is \textbf{bolded}.}
\label{tab:qmsum_relative_to_random_scores}
\end{table}

\begin{table}[H]
\centering
\resizebox{1\linewidth}{!}{
\begin{tabular}{lcccccccccc}
\toprule
& \multicolumn{5}{c}{BART} & \multicolumn{5}{c}{Llama 3} \\
 & R-1 & R-2 & R-L & METEOR & BERTScore & R-1 & R-2 & R-L & METEOR & BERTScore \\
\cmidrule(r){2-6}\cmidrule(r){7-11}

 $\literalsumm$            & $-$\underline{3.82} & $-$\underline{3.14} & $-$\underline{1.44} & $-$\underline{6.06}  & $-$2.73                 & $+$0.74              & $+$1.37              & $+$0.24              & $+$1.20                 & $-$1.65                    \\\cmidrule(r){2-6}\cmidrule(r){7-11}
 $\pragansonly$             & $-$1.37             & $-$1.57             & $-$0.70             & $-$1.49              & $-$\underline{2.80}     & $-$\underline{0.08}  & $-$\underline{0.20}  & $+$0.06              & $-$\underline{0.55}     & $-$\underline{2.64}        \\
 $\pragsourceonly$          & $+$0.34             & $+$\textbf{0.09}    & $+$\textbf{0.00}    & $+$1.57              & $+$\textbf{0.48}        & $+$0.39              & $+$0.25              & $-$\underline{0.02}  & $+$0.23                 & $-$0.51                    \\\cmidrule(r){2-6}\cmidrule(r){7-11}
 \optimalpragans             & $+$\textbf{0.57}    & $-$0.25             & $-$0.27             & $+$\textbf{2.18}     & $-$2.59                 & $+$0.74              & $+$1.37              & $+$\textbf{0.40}     & $+$1.17                 & $+$\textbf{1.35}           \\
 \optimalpragsource         & $-$1.34             & $-$1.22             & $-$0.70             & $-$1.33              & $-$1.86                 & $+$0.78              & $+$\textbf{1.41}     & $+$0.24              & $+$\textbf{1.23}        & $-$0.38                    \\
 \optimalpragas & $+$0.38             & $-$0.02             & $-$0.22             & $+$1.80              & $-$0.74                 & $+$\textbf{0.80}     & $+$1.39              & $+$0.35              & $+$1.23                 & $+$0.96                    \\
\bottomrule
\end{tabular}
}
\caption{Relative change in ROUGE, METEOR, and BERTScore scores between the pragmatic summarizers and $\pragrandom$ on the \squality dataset. Relative changes are aggregated across all decoding methods. A value of $+x$ ($-x$) indicates that the average metric score achieved is $x\%$ higher (lower) than the one achieved by $\pragrandom$. The lowest relative change is \underline{underlined}, and the highest is \textbf{bolded}.}
\label{tab:squality_relative_to_random_scores}
\end{table}

\subsection{Text Quality}

\label{sec:app_text_quality}

\subsubsection{Comparison of Alternative Pragmatic Summarizers with SEAHORSE Metrics}

\label{sec:app_seahorse_metrics_tables}

\begin{table}[H]
\centering
\resizebox{\linewidth}{!}{
\begin{tabular}{lcccccccc}
\toprule
& \multicolumn{4}{c}{BART} & \multicolumn{4}{c}{Llama 3} \\
 & Comp. & Rep. & Gram. & Conc. & Comp. & Rep. & Gram. & Conc. \\
\cmidrule(r){2-5}\cmidrule(r){6-9}

 $\pragrandom$                    & 0.8311              & 0.9505             & 0.6469              & 0.2180              & 0.8889                 & 0.9454                & \textbf{0.8171}        & 0.3664                 \\\cmidrule(r){2-5}\cmidrule(r){6-9}
 $\literalsumm$            & \textbf{0.8931}     & \textbf{0.9680}    & \textbf{0.7620}     & \textbf{0.2497}     & 0.8448                 & 0.9406                & 0.7802                 & \textbf{0.3872}        \\
 $\pragansonly$             & 0.7977              & 0.9453             & 0.6265              & 0.1937              & 0.8635                 & 0.9484                & 0.7898                 & 0.3535                 \\
 $\pragsourceonly$          & 0.7368              & 0.9176             & 0.5616              & 0.1828              & 0.7552                 & 0.9188                & 0.6777                 & 0.2975                 \\\cmidrule(r){2-5}\cmidrule(r){6-9}
 \optimalpragans             & 0.8608              & 0.9585             & 0.7215              & 0.2344              & \textbf{0.9005}        & \textbf{0.9567}       & 0.8058                 & 0.3721                 \\
 \optimalpragsource         & 0.8928              & 0.9679             & 0.7609              & 0.2497              & 0.8451                 & 0.9431                & 0.7809                 & 0.3866                 \\
 \optimalpragas & 0.8891              & 0.9659             & 0.7571              & 0.2470              & 0.8974                 & 0.9564                & 0.8024                 & 0.3759                 \\\cmidrule(r){2-5}\cmidrule(r){6-9}
 $\pragoracle$                    & 0.9613              & 0.9865             & 0.9040              & 0.3715              & 0.9738                 & 0.9841                & 0.9315                 & 0.5405                 \\
\bottomrule
\end{tabular}
}
\caption{Text quality of the summaries select by different pragmatic summarizers on the \multioped dataset. The summaries generated by the literal summarizer used standard sampling (temperature of 2). Text quality is measure via the SEAHORSE metrics Comprehensible (Comp.), Repetition (Rep.), Grammar (Gram.), and Conciseness (Conc.).}
\label{tab:multioped_temp_2}
\end{table}

\begin{table}[H]
\centering
\resizebox{\linewidth}{!}{
\begin{tabular}{lcccccccc}
\toprule
& \multicolumn{4}{c}{BART} & \multicolumn{4}{c}{Llama 3} \\
 & Comp. & Rep. & Gram. & Conc. & Comp. & Rep. & Gram. & Conc. \\
\cmidrule(r){2-5}\cmidrule(r){6-9}

 $\pragrandom$                    & 0.4911              & 0.7299             & \textbf{0.4266}     & 0.0969              & 0.6843                 & 0.8228                & 0.6795                 & 0.2853                 \\\cmidrule(r){2-5}\cmidrule(r){6-9}
 $\literalsumm$            & 0.4901              & 0.7305             & 0.4162              & 0.1024              & 0.6496                 & 0.7839                & 0.6439                 & 0.3045                 \\
 $\pragansonly$             & \textbf{0.4964}     & 0.7228             & 0.4217              & 0.0911              & 0.6794                 & 0.8158                & 0.6769                 & 0.2931                 \\
 $\pragsourceonly$          & 0.4809              & 0.6943             & 0.3996              & 0.0915              & 0.6360                 & 0.7869                & 0.6320                 & 0.2769                 \\\cmidrule(r){2-5}\cmidrule(r){6-9}
 \optimalpragans             & 0.4961              & \textbf{0.7343}    & 0.4234              & \textbf{0.1026}     & \textbf{0.6867}        & \textbf{0.8292}       & \textbf{0.6907}        & 0.3015                 \\
 \optimalpragsource         & 0.4906              & 0.7304             & 0.4165              & 0.1024              & 0.6657                 & 0.8163                & 0.6706                 & \textbf{0.3059}        \\
 \optimalpragas & 0.4923              & 0.7297             & 0.4174              & 0.1022              & 0.6856                 & 0.8279                & 0.6896                 & 0.3047                 \\\cmidrule(r){2-5}\cmidrule(r){6-9}
 $\pragoracle$                    & 0.6897              & 0.8650             & 0.6654              & 0.1760              & 0.7669                 & 0.9131                & 0.7853                 & 0.4424                 \\
\bottomrule
\end{tabular}
}
\caption{Text quality of the summaries select by different pragmatic summarizers on the \qmsum dataset. The summaries generated by the literal summarizer used standard sampling (temperature of 2). Text quality is measure via the SEAHORSE metrics Comprehensible (Comp.), Repetition (Rep.), Grammar (Gram.), and Conciseness (Conc.).}
\label{tab:qmsum_temp_2}
\end{table}

\begin{table}[H]
\centering
\resizebox{\linewidth}{!}{
\begin{tabular}{lcccccccc}
\toprule
& \multicolumn{4}{c}{BART} & \multicolumn{4}{c}{Llama 3} \\
 & Comp. & Rep. & Gram. & Conc. & Comp. & Rep. & Gram. & Conc. \\
\cmidrule(r){2-5}\cmidrule(r){6-9}

 $\pragrandom$                    & 0.3389              & 0.8354             & 0.3269              & \textbf{0.0842}     & 0.7488                 & 0.8673                & 0.6991                 & 0.2354                 \\\cmidrule(r){2-5}\cmidrule(r){6-9}
 $\literalsumm$            & 0.4510              & 0.8711             & 0.3740              & 0.0668              & 0.7701                 & 0.8942                & 0.7475                 & \textbf{0.2968}        \\
 $\pragansonly$             & 0.3943              & 0.8327             & 0.3686              & 0.0798              & 0.7423                 & 0.9028                & 0.6998                 & 0.2414                 \\
 $\pragsourceonly$          & 0.2733              & 0.8020             & 0.2968              & 0.0821              & 0.4316                 & 0.7814                & 0.4241                 & 0.1389                 \\\cmidrule(r){2-5}\cmidrule(r){6-9}
 \optimalpragans             & 0.4464              & 0.8620             & \textbf{0.3835}     & 0.0830              & \textbf{0.7953}        & \textbf{0.9100}       & \textbf{0.7516}        & 0.2941                 \\
 \optimalpragsource         & \textbf{0.4530}     & \textbf{0.8748}    & 0.3751              & 0.0793              & 0.7701                 & 0.8949                & 0.7475                 & \textbf{0.2968}        \\
 \optimalpragas & 0.4517              & 0.8748             & 0.3735              & 0.0808              & 0.7781                 & 0.9018                & 0.7461                 & 0.2917                 \\\cmidrule(r){2-5}\cmidrule(r){6-9}
 $\pragoracle$                    & 0.6832              & 0.9674             & 0.6558              & 0.1610              & 0.9583                 & 0.9831                & 0.9244                 & 0.4335                 \\
\bottomrule
\end{tabular}
}
\caption{Text quality of the summaries select by different pragmatic summarizers on the \squality dataset. The summaries generated by the literal summarizer used standard sampling (temperature of 2). Text quality is measure via the SEAHORSE metrics Comprehensible (Comp.), Repetition (Rep.), Grammar (Gram.), and Conciseness (Conc.).}
\label{tab:squality_temp_2}
\end{table}

\subsubsection{Plots of Summarization Quality and Text Quality}

        \begin{figure}[H]
            \centering
            \includegraphics[width=\linewidth]{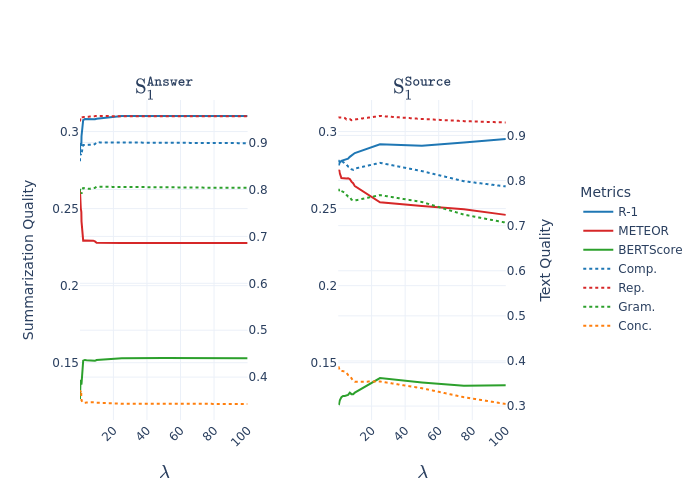}
            \caption{Tradeoff between summarization quality and text quality as controlled by $\lambda$ for the \multioped dataset using Llama 3 with standard sampling (temperature of 2). Solid lines represent summarization quality metrics (ROUGE-1, METEOR, and BERTScore), while dashed lines represent text quality metrics (Comprehensibility, Repetition, Grammar and Conciseness). Note that the left and right y-axes have different scales for summarization quality and text quality, respectively. }
            \label{llama3_qfs_temp_2_multioped_metrics_versus_lambda_complete}
        \end{figure}
        
        \begin{figure}[H]
            \centering
            \includegraphics[width=\linewidth]{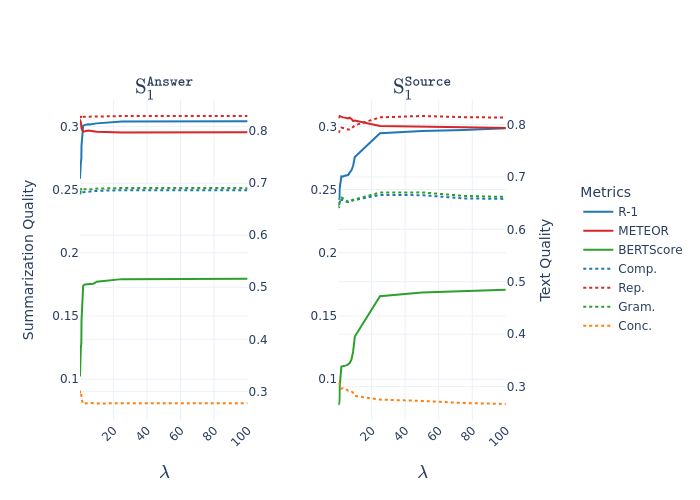}
            \caption{Tradeoff between summarization quality and text quality as controlled by $\lambda$ for the \qmsum dataset using Llama 3 with standard sampling (temperature of 2). Solid lines represent summarization quality metrics (ROUGE-1, METEOR, and BERTScore), while dashed lines represent text quality metrics (Comprehensibility, Repetition, Grammar and Conciseness). Note that the left and right y-axes have different scales for summarization quality and text quality, respectively. }
            \label{llama3_qfs_temp_2_qmsum_metrics_versus_lambda_complete}
        \end{figure}
        
        \begin{figure}[H]
            \centering
            \includegraphics[width=\linewidth]{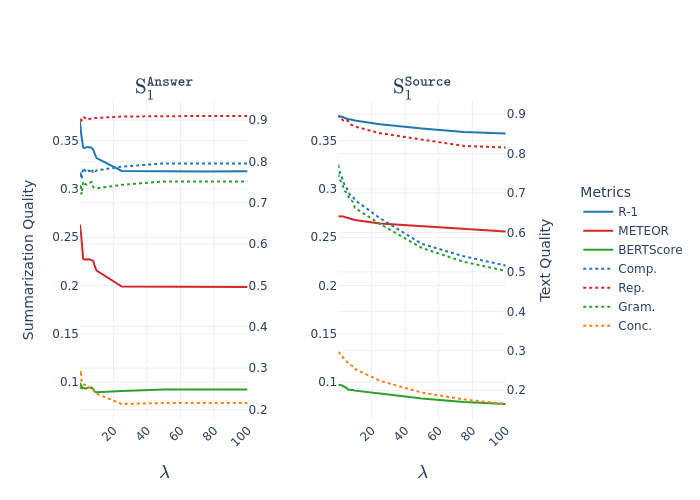}
            \caption{Tradeoff between summarization quality and text quality as controlled by $\lambda$ for the \squality dataset using Llama 3 with standard sampling (temperature of 2). Solid lines represent summarization quality metrics (ROUGE-1, METEOR, and BERTScore), while dashed lines represent text quality metrics (Comprehensibility, Repetition, Grammar and Conciseness). Note that the left and right y-axes have different scales for summarization quality and text quality, respectively. }
            \label{llama3_qfs_temp_2_squality_metrics_versus_lambda_complete}
        \end{figure}

        \begin{figure}[H]
            \centering
            \includegraphics[width=\linewidth]{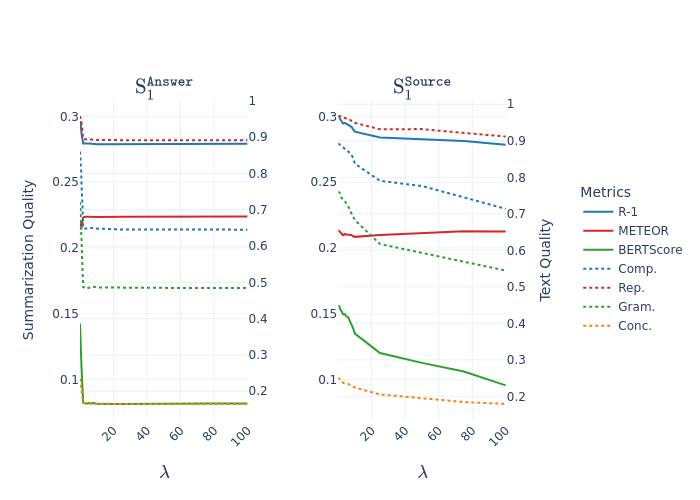}
            \caption{Tradeoff between summarization quality and text quality as controlled by $\lambda$ for the \multioped dataset using BART with standard sampling (temperature of 2). Solid lines represent summarization quality metrics (ROUGE-1, METEOR, and BERTScore), while dashed lines represent text quality metrics (Comprehensibility, Repetition, Grammar and Conciseness). Note that the left and right y-axes have different scales for summarization quality and text quality, respectively. }
            \label{bart_finetuned_qfs_temp_2_multioped_metrics_versus_lambda_complete}
        \end{figure}
        
        \begin{figure}[H]
            \centering
            \includegraphics[width=\linewidth]{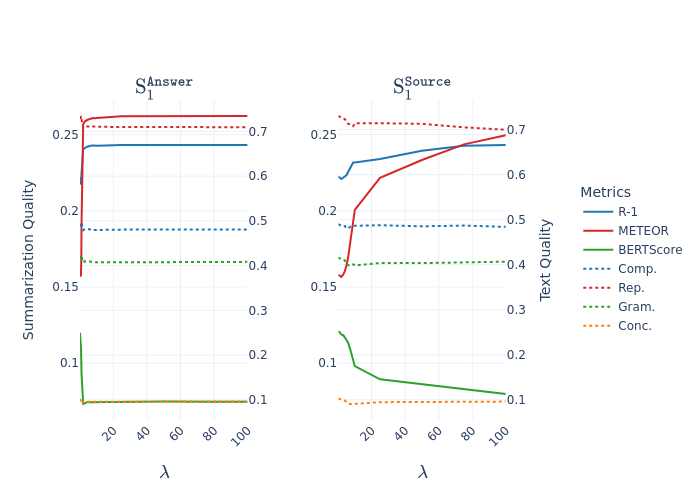}
            \caption{Tradeoff between summarization quality and text quality as controlled by $\lambda$ for the \qmsum dataset using BART with standard sampling (temperature of 2). Solid lines represent summarization quality metrics (ROUGE-1, METEOR, and BERTScore), while dashed lines represent text quality metrics (Comprehensibility, Repetition, Grammar and Conciseness). Note that the left and right y-axes have different scales for summarization quality and text quality, respectively. }
            \label{bart_finetuned_qfs_temp_2_qmsum_metrics_versus_lambda_complete}
        \end{figure}
        
        \begin{figure}[H]
            \centering
            \includegraphics[width=\linewidth]{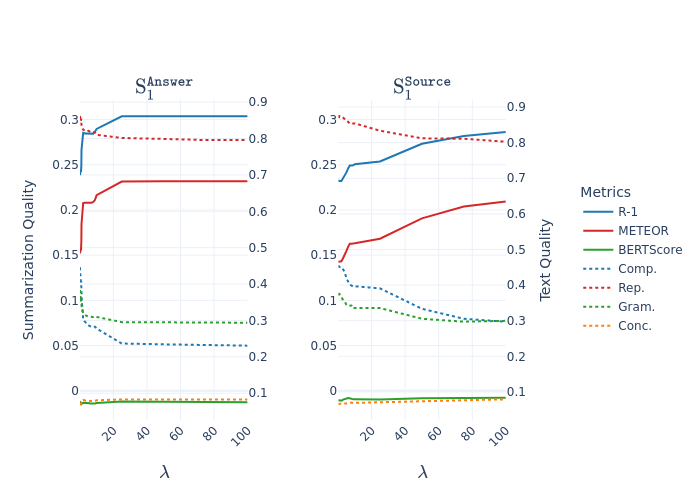}
            \caption{Tradeoff between summarization quality and text quality as controlled by $\lambda$ for the \squality dataset using BART with standard sampling (temperature of 2). Solid lines represent summarization quality metrics (ROUGE-1, METEOR, and BERTScore), while dashed lines represent text quality metrics (Comprehensibility, Repetition, Grammar and Conciseness). Note that the left and right y-axes have different scales for summarization quality and text quality, respectively. }
            \label{bart_finetuned_qfs_temp_2_squality_metrics_versus_lambda_complete}
        \end{figure}

\end{document}